\pdfoutput=1
\documentclass{article}

\usepackage{xcolor} %

\usepackage{microtype}
\usepackage{graphicx}
\usepackage{subfigure}
\usepackage{booktabs} %
\makeatletter
\let\main@addtocurcol\@addtocurcol
\usepackage{flafter}
\let\appx@addtocurcol\@addtocurcol
\let\@addtocurcol\main@addtocurcol
\makeatother

\usepackage{hyperref}
\usepackage{xurl}

\usepackage[accepted]{mlsys2025}
\makeatletter
\renewcommand{\Notice@String}{Preprint.}
\makeatother
\hypersetup{pdfsubject={}}

\usepackage{amsmath}
\usepackage{amssymb}
\usepackage{xspace}
\usepackage{multirow}
\usepackage{pifont} %
\usepackage{array}
\newcommand{\recsa}{\mbox{RecSA}\xspace}

\mlsystitlerunning{Recursive Sparse Attention}

\begin{document}

\twocolumn[
\mlsystitle{Recursive Sparse Attention: Million-Token Length Extrapolation at Logarithmic Cost}

\mlsyssetsymbol{equal}{*}

\begin{mlsysauthorlist}
\mlsysauthor{Ruitian Wang}{equal,ucsd}
\mlsysauthor{Xinyu Yang}{equal,cmu}
\mlsysauthor{Minshen Zhang}{equal,ucsd}
\mlsysauthor{Hao Zhang}{ucsd}
\end{mlsysauthorlist}

\mlsysaffiliation{ucsd}{University of California, San Diego}
\mlsysaffiliation{cmu}{Carnegie Mellon University}

\mlsyscorrespondingauthor{Hao Zhang}{haozhang@ucsd.edu}

\mlsyskeywords{Sparse Attention, Long Context, Length Extrapolation, LLM Inference}

\vskip 0.3in

\begin{abstract}
At context lengths of about 1M tokens, dense attention is costly in prefill and decoding, and training typically
relies on long documents, which become scarcer as the target length increases. Most sparse attention methods score all preceding blocks or tokens, so their
selection cost per query is $O(L)$ at context length~$L$; hierarchical selection reduces it to $O(\log L)$, but
prior trainable hierarchical methods report accuracy only within their training length. We propose \recsa (Recursive Sparse
Attention), which builds a tree over the KV cache in which each node
summarizes its subtree with one key and one mean-pooled value. The keys come from one pooler per layer, shared
by all levels and distilled from dense attention. Each level above the tokens selects at most $k$ candidates, so each query has $O(\log L)$ selection cost and reads $O(\log L)$ entries; the summaries of unselected candidates
share one softmax with the exact tokens. Each entry's RoPE position is its index among the candidates of its level,
so at any length every relative distance stays within the range seen
during training. Initialized from
Qwen3-4B and trained on 0.62B tokens at the original model's 32K training length, \recsa outperforms all baselines on RULER
at every length from 131K to 2M; at 1M, it scores 60.5, compared with 40.0 for the best baseline,
Dense~+~DCA. With our optimized kernels, \recsa is faster than dense attention in end-to-end prefill and decoding from
64K to 2M; at 524K on one H200 GPU, it accelerates prefill by $3.2\times$ and decoding by $3.8\times$.
\end{abstract}
]

\printAffiliationsAndNotice{\mlsysEqualContribution}

\section{Introduction}
\label{sec:intro}

State-of-the-art language models now support context windows of about 1M
tokens~\citep{deepseek2026v4}. At such lengths, inference is costly and training data is scarce. In prefill, the
computation of dense attention grows quadratically with length, and in decoding, every generated token reads the
entire KV cache. Training at the target length relies on sufficiently long documents, which become scarcer as the
length increases~\citep{gao2025nextlong}. Some of these models reduce the inference cost with sparse
attention~\citep{deepseek2026v4}. As language models continue to develop, the context they need is likely to keep
growing. Longer contexts call for methods whose inference cost grows slowly with length and that need no training
at the target length.

\begin{figure}[!t]
\centering
 \includegraphics[width=0.98\columnwidth]{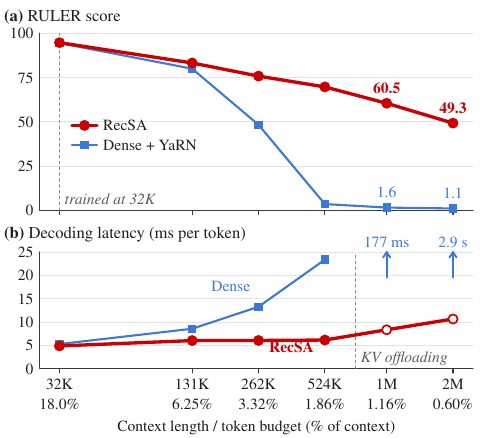}
\caption{RULER score and decoding latency from 32K to 2M. (b)~End-to-end latency on one H200 GPU with batch size~1; Dense uses FlashAttention in the same framework. At 1M and 2M, host memory holds the exact tokens of \recsa (\S\ref{sec:offload}) and the part of the Dense KV cache beyond GPU memory; hollow markers indicate offloading. Arrows mark off-scale values. Percentages show the token budget of \recsa as a fraction of the context length.}
\label{fig:teaser}
\end{figure}

Trainable sparse attention methods such as MoBA~\citep{lu2025moba}, NSA~\citep{yuan2025nsa} and
DSA~\citep{deepseek2025v32} are trained at the target length. HSA~\citep{hu2025hsa,leng2026hsalength,hu2026hsaultralong}
and HiLS~\citep{hu2026hils} train only on short sequences, and
\mbox{InfLLM}~\citep{xiao2024infllm} and InfiniteHiP~\citep{lee2025infinitehip} are training-free; neither group needs
long documents to run far beyond the training length. All of these methods score every preceding block or token at the
first level, so at context length~$L$ the selection cost per query is $O(L)$. Hierarchical selection scores a top level
of bounded size and then only the children of selected nodes, which reduces this cost to $O(\log L)$.
HiP~\citep{lee2025hip} and PISA~\citep{tang2026pisa} keep the original position of each key, so beyond the
training length the distances to selected keys can exceed those seen in training. Hierarchical selection alone does
not remove the need for context extension.

We propose \recsa, a Recursive Sparse Attention mechanism with hierarchical selection. It groups every 16 consecutive KV
entries under a parent node and applies the same grouping to parent nodes, building a tree from the bottom up. Each
node has one key and one mean-pooled value, which form the \emph{summary} of its whole subtree. Selection starts at
the top level, whose nodes form the first \emph{candidate list}, and at each level scores every candidate against the query and selects at most $k$ of them. Their children form the candidate list of the next level, and this continues until the candidates are \emph{exact tokens}. Unselected candidates also
participate in the attention computation: their summaries and all exact tokens in the last candidate list share one
softmax. Whenever the top level exceeds a fixed number of nodes, a level is added above it, so every candidate list has a bounded size, and the selection cost and the number of entries read per query are both $O(\log L)$.
RoPE~\citep{su2024roformer} is applied to every entry, summaries included, and the position of an entry is its index in
the candidate list of its level, with the query at the last index. Every relative distance to the query is thus below 4{,}096, the maximum size of a candidate list with the default $k = 256$. Longer sequences introduce no distance unseen during training at 32K.

A learnable \emph{pooler} computes the key of a node from the keys of its children. Each attention layer has one
pooler, applied recursively at every level of the tree, so adding a level requires no new parameters. Training starts
from a pretrained dense model and has two stages. The first stage freezes the backbone and aligns the softmax over the scores of the candidate nodes at each level with the distribution of dense attention
mass over their subtrees. The second stage freezes the pooler and trains the backbone under \recsa to adapt it to sparse
attention. Both stages use only sequences at the training length of the original model.

We evaluate \recsa on Qwen3-4B~\citep{qwen2025qwen3}, whose training length is 32K. The two stages process 0.62B tokens in
total. For comparison, we train dense and sparse backbones on the same data and evaluate every backbone with
each of two context extension methods, YaRN~\citep{peng2024yarn} and DCA~\citep{an2024dca}; we also compare with two
training-free methods. At every length, the \emph{token budget} of
each sparse method, i.e., the number of entries a query attends to, is matched to that of \recsa. The evaluation covers
synthetic long-context tasks, perplexity on long texts, real-world long-document tasks and short-context benchmarks.
At 1M, the token budget of \recsa is only 1.16\% of the context, and \recsa scores 60.5 on RULER~\citep{hsieh2024ruler},
compared with 1.6 for Dense~+~YaRN, as shown in \mbox{Figure}~\ref{fig:teaser}a. The strongest of the other baselines,
Dense~+~DCA, scores 40.0 (Table~\ref{tab:ruler}). At 32K, \recsa matches Dense~+~YaRN on RULER; on
short-context benchmarks, \recsa performs comparably to the original model.

We implement prefill and decoding for \recsa in Triton~\citep{tillet2019triton} and CUDA. In prefill, some existing
block-sparse kernels group adjacent queries and load the union of their selected blocks once for the
group~\citep{hu2026hils}. This sharing reuses each key, already rotated by RoPE, for every query in the group, but
in \recsa the index of a key, and hence its rotation, varies with the candidate list of each query. Because RoPE makes the attention score depend on positions only through the relative position, we split the rotation between keys and queries so that the part applied to each key is the same for all queries, and the group still loads each KV block once.
On one H200 GPU with batch
size~1, \recsa accelerates prefill by $3.2\times$ and decoding by $3.8\times$ at 524K, compared with dense
FlashAttention~\citep{dao2024flashattention2} in the same framework, and its decoding latency stays nearly constant
from 131K to 524K (Figure~\ref{fig:teaser}b).

We make the following contributions:
\begin{itemize}
\item We propose \recsa, which combines hierarchical selection over a tree built on the KV cache with summaries for the
unselected candidates and candidate-local RoPE (\S\ref{sec:tree} and \S\ref{sec:rope}).
\item We distill the pooler, shared across levels, from dense attention and train \recsa only on sequences of
32K tokens (\S\ref{sec:pooler} and \S\ref{sec:training}).
\item We provide prefill and decoding kernels for \recsa; in prefill, a group of adjacent queries loads each KV block once,
although their candidate lists give its keys different positions (\S\ref{sec:system}).
\item On Qwen3-4B, we compare \recsa from 32K to 2M with dense and sparse baselines, trained on the same data or
training-free, at matched token budgets (\S\ref{sec:eval}).
\end{itemize}

\section{Background}
\label{sec:background}

\paragraph{Sparse attention.}
Sparse attention methods differ in how they select the entries each query attends to. Block-level methods select whole
blocks of consecutive tokens. MoBA~\citep{lu2025moba} scores each block by the dot product of the query with the mean of
the block's keys, and the query attends to the tokens of the highest-scoring blocks and of the current block, which
contains the query. NSA~\citep{yuan2025nsa} gates three attention branches: over compressed blocks, over selected blocks
and over a sliding window. Token-level methods select individual tokens:
in DSA~\citep{deepseek2025v32}, a lightning indexer with a few heads scores every preceding token, and the query attends
to the highest-scoring tokens. We count every entry a query attends to toward the token budget,
including entries that summarize several tokens and the tokens of attention sinks, local windows and the current block.

\paragraph{Length extrapolation.} Positional encoding plays an important role in how LLMs generalize beyond their training length. Transformers without explicit positional encodings (NoPE) can generalize better~\citep{kazemnejad2023}, but remain limited by increasing attention dispersion~\citep{wang2024lengthgeneralizationcausaltransformers}. Most LLMs, however, use RoPE~\citep{su2024roformer}, which rotates a query at position $t$ by $R_t$ and a key at position $j$ by $R_j$, so scores depend only on $t - j$. Consequently, extrapolation introduces out-of-distribution relative distances~\citep{an2024dca}. YaRN~\citep{peng2024yarn} mitigates this by rescaling the rotation angles of the low-frequency
dimensions, while DCA~\citep{an2024dca} partitions the sequence into shorter chunks and remaps positions to stay within the training range. Sparse attention also improves extrapolation by limiting attention to relevant parts of a longer context~\citep{xiao2024infllm,leng2026hsalength}.

\section{Recursive Sparse Attention}
\label{sec:recsa}

\begin{figure*}[!t]
\centering
\includegraphics[width=\textwidth]{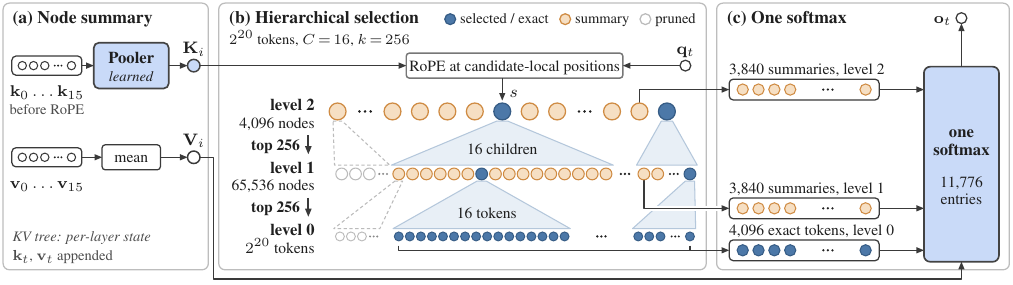}\par
\caption{One attention layer of \recsa ($2^{20}$ tokens, $C = 16$, $k = 256$, three levels). (a) The pooler computes the key $\mathbf{K}_i$ of a node from the keys of its 16 children; its value
$\mathbf{V}_i$ is the mean of their values. (b) $s$ is the score of each candidate for the query $\mathbf{q}_t$ (Eq.~\eqref{eq:score});
levels 2 and 1 each select 256 candidates by this score, whose children form the next candidate list, and level 0 has no
selection; pruned nodes are not candidates. (c) Summaries
and exact tokens share one softmax, with output $\mathbf{o}_t$.}
\label{fig:arch}
\end{figure*}

To scale context length without requiring longer training data or incurring quadratic attention costs, we introduce Recursive Sparse Attention (RecSA), as shown in Figure~\ref{fig:arch}. In our model,
each KV head is shared by a group $\mathcal{G}$ of $G = 4$ query heads through grouped-query attention
(GQA)~\citep{ainslie2023gqa}. We describe one such group for the query at position $t$, using zero-based indices.

\subsection{Tree Construction and Hierarchical Selection}
\label{sec:tree}

\paragraph{Tree construction.}
Each attention layer maintains one KV tree per KV head. Level 0 holds the key $\mathbf{k}_j$ of each token
$j$, before RoPE (\S\ref{sec:rope}), and its value $\mathbf{v}_j$. Each group of $C = 16$ consecutive entries at level
$\ell - 1$ forms one parent node at level $\ell$, up to the top level $D - 1$ ($D$ levels in total). The pooler
$\mathrm{Pool}$ computes the key of each node (\S\ref{sec:pooler}), and the value of a node is the mean of its children's
values:
\begin{equation}
\label{eq:tree}
\begin{aligned}
\mathbf{K}^{(\ell)}_i &= \mathrm{Pool}\big(\mathbf{K}^{(\ell-1)}_{Ci}, \dots, \mathbf{K}^{(\ell-1)}_{Ci+C-1}\big), \\
\mathbf{V}^{(\ell)}_i &= \frac{1}{C} \sum_{c=0}^{C-1} \mathbf{V}^{(\ell-1)}_{Ci+c},
\end{aligned}
\end{equation}
where $\mathbf{K}^{(\ell)}_i$ and $\mathbf{V}^{(\ell)}_i$ are the key and value of node $i$ at level $\ell$, with
$\mathbf{K}^{(0)}_j = \mathbf{k}_j$ and $\mathbf{V}^{(0)}_j = \mathbf{v}_j$. The key and value of a node form the
summary of its subtree.

\paragraph{Hierarchical selection.}
Selection proceeds from the top level down (Figure~\ref{fig:arch}b). Let $a^{(\ell)}_t = \lfloor t / C^{\ell} \rfloor$
denote the node of level $\ell$ that contains position $t$, which we call the \emph{current node}. The candidate list
$\mathcal{A}^{(\ell)}_t$ of level $\ell$ is ordered by position in the sequence; at the top level, it holds the current
node and all nodes before it. Query head $h$ scores candidate $i$ with its query $\mathbf{q}_{t,h}$:
\begin{equation}
\label{eq:score}
\begin{aligned}
s^{(\ell)}_{t,h,i} &= \big(R_{n_\ell - 1}\, \mathbf{q}_{t,h}\big)^{\top} \big(R_{p_i} \mathbf{K}^{(\ell)}_i\big) / \sqrt{d}, \\
I^{(\ell)}_{t,i} &= \textstyle\sum_{h \in \mathcal{G}} \exp(s^{(\ell)}_{t,h,i}),
\end{aligned}
\end{equation}
where $d = 128$ is the head dimension, $R_p$ is the RoPE rotation at position $p$, $p_i$ is the index of candidate $i$
in $\mathcal{A}^{(\ell)}_t$ and $n_\ell = |\mathcal{A}^{(\ell)}_t|$, so the query takes the last index
(\S\ref{sec:rope}). As in NSA~\citep{yuan2025nsa}, the query heads in $\mathcal{G}$ share one selection, based on the
importance $I^{(\ell)}_{t,i}$. At each level $\ell \ge 1$, \recsa selects the current node and the $k - 1$ other
candidates with the highest importance ($k = 256$ by default), and the children of the selected nodes, up to the query,
form the candidate list of the next level:
\begin{equation}
\label{eq:select}
\begin{aligned}
\mathcal{A}^{(D-1)}_t &= [0, \dots, a^{(D-1)}_t], \\
\mathcal{S}^{(\ell)}_t &= \{a^{(\ell)}_t\} \cup \mathrm{TopK}_{k-1}\big(\{I^{(\ell)}_{t,i}\}_{i \in \mathcal{A}^{(\ell)}_t \setminus \{a^{(\ell)}_t\}}\big), \\
\mathcal{A}^{(\ell-1)}_t &= \mathop{\mathrm{concat}}\limits_{i \in \mathcal{S}^{(\ell)}_t} [Ci, \dots, \min(Ci{+}C{-}1, a^{(\ell-1)}_t)],
\end{aligned}
\end{equation}
where $\ell = D - 1, \dots, 1$, $\mathrm{TopK}_{k-1}$ returns the indices of the $k - 1$ candidates with the highest
importance and $\mathrm{concat}$ joins the child lists in increasing order of $i$, so $|\mathcal{A}^{(\ell-1)}_t| \le
Ck$. Level 0 has no top-$k$ selection: the query attends to all of its candidate tokens (Eq.~\eqref{eq:softmax}). Among
the candidates of each level $\ell \ge 1$, only the current node can contain tokens after the query. This node takes
no part in the ranking and is always expanded, with its children taken only up to the query. Hence all tokens of the level-1 current node up to the query enter the softmax as exact tokens, and every summary covers a complete subtree before
the query. \recsa adds no attention sink~\citep{xiao2024streamingllm} or fixed local window.

\paragraph{Depth.}
For the query at position $t$, the number of levels $D$ is set by its prefix, positions 0 to $t$: whenever the
top level exceeds $Ck$ nodes, a level is added above it. If the prefix has at most $Ck$ tokens, $D = 1$ and \recsa
is equivalent to dense causal attention. With
the default $k = 256$, $Ck = 4{,}096$; beyond 4{,}096 tokens, a prefix has two levels up to 65{,}536 tokens,
three up to 1{,}048{,}576 tokens and four at 2M.

\paragraph{One softmax over summaries and tokens.}
The summaries of the unselected candidates and the exact tokens of level 0 share one softmax (Figure~\ref{fig:arch}c),
over entries $e = (\ell, i)$ identified by level and index:
\begin{equation}
\label{eq:softmax}
\begin{aligned}
\mathcal{E}_t &= (\{0\} \times \mathcal{A}^{(0)}_t) \cup \textstyle\bigcup_{\ell=1}^{D-1} (\{\ell\} \times (\mathcal{A}^{(\ell)}_t \setminus \mathcal{S}^{(\ell)}_t)), \\
\mathbf{o}_{t,h} &= \frac{\sum_{e \in \mathcal{E}_t} \exp(s_{t,h,e})\, \mathbf{V}_e}{\sum_{e \in \mathcal{E}_t} \exp(s_{t,h,e})},
\end{aligned}
\end{equation}
where, for $e = (\ell, i)$, $s_{t,h,e} = s^{(\ell)}_{t,h,i}$ is the score of Eq.~\eqref{eq:score} and $\mathbf{V}_e =
\mathbf{V}^{(\ell)}_i$, and $\mathbf{o}_{t,h}$ is the output of head $h$. Because the top-level candidates cover the prefix and each selected node is replaced by its children up to the query, the summaries and the exact tokens partition the prefix. The softmax uses the score of a summary as its logit, with no correction for subtree size or level.

\paragraph{Cost.}
For fixed $C$ and $k$, each level scores at most $Ck$ candidates, and the number of levels grows logarithmically with
$L$, so the selection cost per query is $O(\log L)$. The token budget is the number of entries a query attends to,
$|\mathcal{E}_t| \le Ck + (D - 1)(Ck - k)$: at most $Ck$ exact tokens at level 0 and, at each level above, the summaries
of at most $Ck - k$ unselected candidates (3{,}840 with the default $k$). The number of entries a query
reads, counting each candidate once, is the token budget plus the keys of the $(D - 1)k$ selected nodes, which are
expanded instead of entering the softmax; it is at most $D \cdot Ck$, also $O(\log L)$.

\subsection{Candidate-Local RoPE}
\label{sec:rope}

\begin{figure}[!t]
\centering
\includegraphics[width=\columnwidth]{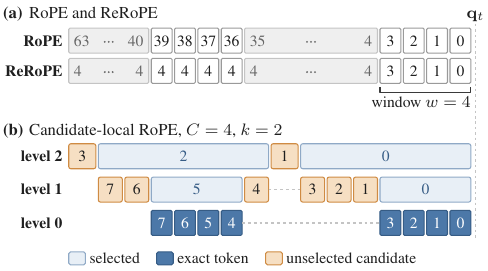}\par
\caption{Relative distance to the query $\mathbf{q}_t$, the last of 64 tokens. (a) RoPE and ReRoPE~\citep{su2023rerope};
gray bars stand for runs of tokens. (b) Candidate-local RoPE on an example tree with $C = 4$ children per node and
$k = 2$ selected nodes per level: each selected node is drawn above its children, and at every level the distances stay below $Ck = 8$. Dashed spans mark parts of the sequence skipped
between candidates; they receive no positions.}
\label{fig:rope}
\end{figure}

\recsa applies RoPE~\citep{su2024roformer} with one change, which we call \emph{candidate-local RoPE}: every candidate
of level $\ell$, summaries included, takes its index in $\mathcal{A}^{(\ell)}_t$ as its position, and the query takes
the last index, that of the current node. Parts of the sequence skipped between candidates receive no
positions. Each level has at most
$Ck = 4{,}096$ candidates (\S\ref{sec:tree}), so the relative distance between the query and any candidate always lies between 0 and
4{,}095. During backbone training on 32K sequences with $k = 256$ (\S\ref{sec:training}), the candidate list of level 0
reaches 4{,}096 tokens, so training covers every relative distance that occurs at inference with the same $k$.
Figure~\ref{fig:rope} compares candidate-local RoPE with ReRoPE~\citep{su2023rerope}, which puts every token beyond
a window $w$ at distance $w$ from the query; under candidate-local RoPE, the candidates of one level are all at different distances from the query.

Because one node can have different indices in the candidate lists of different queries, the tree stores keys before
RoPE, and the pooler (\S\ref{sec:pooler}) pools these keys without position input. We keep the RoPE base $\theta = 10^6$ of the original model,
without scaling.

\subsection{Pooler Distillation}
\label{sec:pooler}

Because the key of a node is used both in the importance shared by a GQA group (Eq.~\eqref{eq:score}) and
in the logit of each head in the final softmax (Eq.~\eqref{eq:softmax}), both should reflect the dense attention
mass on the subtree of the node, so we supervise the pooler with this mass. A mean-pooled key generally does not represent this mass
accurately~\citep{hu2025hsa,leng2026hsalength}. The pooler computes the key $\mathbf{K}$ of node $i$ at level $\ell$
from the keys of its children, $\mathbf{x}_c = \mathbf{K}^{(\ell-1)}_{Ci+c}$ for $c = 0, \dots, C - 1$
(Figure~\ref{fig:arch}a):
\begin{equation}
\label{eq:pooler}
\begin{aligned}
\mathbf{K} &= \boldsymbol{\mu} + g\,(\mathbf{p} - \boldsymbol{\mu}) + \sigma \mathbf{m}, \\
\alpha_c &= \mathrm{softmax}_c\big(\tau\, \mathbf{w}^{\top} (\mathbf{x}_c - \boldsymbol{\mu}) / \sigma\big),\;
\mathbf{p} = \textstyle\sum_c \alpha_c \mathbf{x}_c, \\
\mathbf{m} &= W_2\, \mathrm{SiLU}\big(W_1 [\boldsymbol{\mu} / \rho;\, (\mathbf{p} - \boldsymbol{\mu}) / \sigma] + \mathbf{b}_1\big),
\end{aligned}
\end{equation}
where $\boldsymbol{\mu}$ is the mean of the $\mathbf{x}_c$, $\sigma$ is the root mean square of the elements of all
$\mathbf{x}_c - \boldsymbol{\mu}$, $\rho$ is the root mean square of the elements of $\boldsymbol{\mu}$,
$\mathbf{w}$ is a scoring vector, the scalars $\tau$ and $g$ are an inverse temperature and a gate, $[\cdot\,;\cdot]$
denotes concatenation and $\sigma \mathbf{m}$ is the MLP term. Mean pooling is the special case $g = 0$, $W_2 = 0$. Each attention layer has one pooler with 24{,}770
parameters, shared across levels and across all KV heads of the layer. Appendix~A gives the parameter shapes and implementation details.

For each level $\ell \ge 1$, we compute the attention mass that the frozen original model, with RoPE at the original
positions, puts on the subtree of each candidate in $\mathcal{A}^{(\ell)}_t$, and average it over the $G$ heads of the
group. Below the top level, the candidates cover only part of the prefix, so their masses need not sum to one; normalizing
them over the candidate list gives the teacher distribution $\mathbf{y}^{(\ell)}_t$. The student
distribution $\boldsymbol{\pi}^{(\ell)}_t$ normalizes the importance
of Eq.~\eqref{eq:score} over the same list. Both distributions use the queries and keys of the same forward pass of
the frozen original model, so the loss of each layer depends only on the pooler of that layer. We fit the student distribution to the teacher distribution with a KL
divergence:
\begin{equation}
\label{eq:kl}
\begin{aligned}
\pi^{(\ell)}_{t,i} &= I^{(\ell)}_{t,i} \big/ \textstyle\sum_{i' \in \mathcal{A}^{(\ell)}_t} I^{(\ell)}_{t,i'}, \\
\mathcal{L}_{\mathrm{KL}} &= \textstyle\sum_{t} \sum_{\ell=1}^{D-1} \mathrm{KL}\big(\mathbf{y}^{(\ell)}_t \,\Vert\, \boldsymbol{\pi}^{(\ell)}_t\big).
\end{aligned}
\end{equation}
Below the top level, the candidate lists come from the student's own
selection, which follows the same rule as at inference; no gradient flows through the selection.

The MLP term of the pooler can shift the keys of all nodes of a level by the same vector $\boldsymbol{\delta}$. Under
candidate-local RoPE, the score of candidate $i$ then gains
$\mathbf{q}_{t,h}^{\top} R_{p_i - n_\ell + 1}\, \boldsymbol{\delta} / \sqrt{d}$, which depends on the query and on the
position of the candidate, but not on the content of the candidate's subtree. Training can lower the KL through
this positional bias, without the pooler learning to distinguish subtrees by their content. We add a \emph{shift penalty}
on the mean of the MLP term over the $M_\ell$ nodes of each level $\ell$ of a tree, that is, on the shift shared by
these nodes, but not on the differences between them:
\begin{equation}
\label{eq:shift}
\mathcal{L}_{\mathrm{shift}} = \sum_{\ell=1}^{D-1}
\Big\Vert \frac{1}{M_\ell} \sum_{i} \sigma_i \mathbf{m}_i \Big\Vert^2 \Big/ \Big(\frac{1}{M_\ell} \sum_{i} \Vert \boldsymbol{\mu}_i \Vert^2\Big),
\end{equation}
where $\sigma_i$, $\mathbf{m}_i$ and $\boldsymbol{\mu}_i$ are those of node $i$
in Eq.~\eqref{eq:pooler}; the denominator makes the penalty independent of the key scale.

\subsection{Training}
\label{sec:training}

\paragraph{Stage 1: Pooler.}
We freeze the backbone and train only the parameters of the pooler (\S\ref{sec:pooler}). Because the pooler
(Eq.~\eqref{eq:pooler}) does not involve $k$, this stage uses a smaller $k = 64$ to obtain deeper trees from 32K
sequences. With $Ck = 1{,}024$, the training sequences of 32K tokens have three levels, and the pooler learns to
construct the keys of level 2 from the pooled keys of level 1. Inference with the default $k = 256$ uses at most these three levels up to 1M; at 2M, it adds level
3, which reuses the same parameters. Stage~1 has two parts on the same sequences. The first part
trains $\mathbf{w}$, $\tau$ and $g$ with the MLP term off ($W_2$ fixed at 0). Its loss is $\mathcal{L}_{\mathrm{KL}}$; on retrieval and question-answering sequences with a single piece of evidence, we add an auxiliary cross-entropy whose target is the top-level node that contains the evidence. The second part freezes $\mathbf{w}$, $\tau$ and $g$ and trains $W_1$,
$\mathbf{b}_1$ and $W_2$ with $\mathcal{L}_{\mathrm{KL}} + \lambda \mathcal{L}_{\mathrm{shift}}$.
Appendix~E gives the hyperparameters and the sampled query positions.

\paragraph{Stage 2: Backbone.}
We freeze the pooler and train the whole backbone under \recsa with the default $k = 256$
and only the language modeling loss (data in \S\ref{sec:setup}). Gradients also reach the backbone through the keys and values of the summaries, via the frozen pooler and mean
pooling, respectively; no gradient flows through the selection.

\section{Efficient Prefill and Decoding}
\label{sec:system}

\begin{figure*}[!t]
\centering
\includegraphics[width=\textwidth]{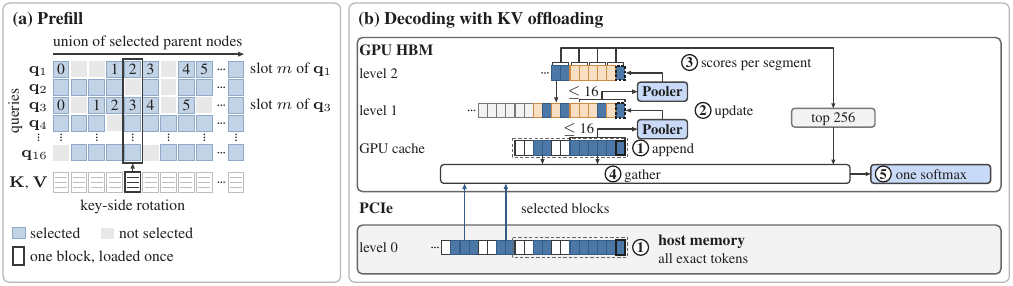}\par
\caption{Prefill and decoding of \recsa. (a)~Prefill: rows correspond to the 16 adjacent queries of one thread block
and columns to the union of the parent nodes they selected; numbers give the slot $m$ of the parent node for the query
of the row. The black frame marks one block (the 16 children of a parent node), loaded once per pass and shared by the
queries that selected its parent. (b)~Decoding with KV offloading: blue indicates selected nodes and blocks, light orange unselected candidates
(summaries) and gray other nodes.}
\label{fig:system}
\end{figure*}

\subsection{Prefill}
\label{sec:prefill}

In prefill, adjacent queries select overlapping sets of parent nodes. Some existing block-sparse kernels process a
group of adjacent queries over the union of their selected blocks and load each block once for the group, with its keys
already rotated by RoPE~\citep{hu2026hils}. Under candidate-local RoPE, however, a key can take different positions, and
hence different rotations, in the candidate lists of different queries.

We split the rotation between keys and queries using slots: the $m$-th node that a query selects at level $\ell$, in index
order, is its \emph{slot}~$m$. In Figure~\ref{fig:system}a, the parent node of the framed block is slot 2 of $\mathbf{q}_1$ and
slot 3 of $\mathbf{q}_3$. Eq.~\eqref{eq:select} concatenates the child lists in slot order, and only the last list
(that of the current node) can be truncated, so child $c$ of slot $m$ has index $mC + c$ in
$\mathcal{A}^{(\ell-1)}_t$. Because $R_a^{\top} R_b = R_{b-a}$, the product in Eq.~\eqref{eq:score} for this child
becomes
\begin{equation}
\label{eq:split}
(R_{n-1} \mathbf{q}_{t,h})^{\top} R_{mC+c} \mathbf{K}
= (R_{n-1-mC} \mathbf{q}_{t,h})^{\top} R_{c} \mathbf{K},
\end{equation}
where $n = n_{\ell-1}$ and $\mathbf{K}$ is the key of the child. The key-side rotation $R_c$ depends only on $c$, not
on the query, so the rotated keys of each level are shared by all queries. The rest of the rotation moves to the query
side: each query starts from $R_{n-1}\mathbf{q}_{t,h}$ and is rotated by $R_{-C}$ after each node it selected. Our
kernels give the same output as rotating the keys for each query, within floating-point error.

One thread block processes the 4 query heads of one KV head for 16 adjacent queries (Figure~\ref{fig:system}a). A pass
loops through the union of the parent nodes that the 16 queries selected, in index order: for each parent, the thread
block loads the 16 children into shared memory once and scores them on tensor cores for all 64 query--head pairs,
masking the queries that did not select this parent. On PG-19, each KV block loaded in the pass of level 0 serves 7.5, 5.2 and 4.1
of the 16 queries on average at 32K, 131K and 524K.

Each level $\ell \ge 1$ takes two passes: the first computes the importance of Eq.~\eqref{eq:score}, and after
selection, the second accumulates only the unselected candidates into an online softmax~\citep{dao2024flashattention2}
as summaries. At the top level, whose candidates are the current node and all nodes before it, both passes are dense
and tiled. Level 0 takes one pass, which accumulates all of its candidate tokens into the same softmax.

\subsection{Decoding}
\label{sec:decoding}

In decoding (Figure~\ref{fig:system}b), the key and value of each new token are appended to level 0 (\ding{172}). Except at a step
that adds a level (\S\ref{sec:tree}), only the last node of each level $\ell \ge 1$ changes, and \recsa recomputes its key
and value (\ding{173}) from its 16 or fewer children by Eq.~\eqref{eq:tree}.

With batch size~1, each step has one query, and scoring a level with one thread block for each of the 8 KV heads would
occupy at most 8 of the 132 SMs of an H200. \recsa instead splits the candidate list of each level
into 16 segments, so that at least 128 thread blocks compute the importance, each relative to the maximum score in
its segment (\ding{174}). A second kernel rescales these values to the maximum over all segments and selects the same
top-$k$ as without segments. The softmax accumulation is split into 16 segments in the same way, and their partial
results are merged (\ding{176}).

Because each level has at most $Ck$ candidates (\S\ref{sec:tree}), every kernel can use a fixed launch grid and read the
current context length from GPU memory, so the decoding step of all 36 layers is captured as one CUDA graph and replayed at every
step that does not add a level.

\subsection{KV Offloading}
\label{sec:offload}

At 1M and 2M, the full KV cache takes 137 and 275~GiB, which together with the 7.5~GiB of model weights exceeds the 140~GiB of memory on one H200. \recsa keeps
the exact tokens in pinned host memory, and the GPU needs to hold only the levels above level 0 (less than 7\% of the
full KV cache) and buffers for the gathered exact tokens.

Selection reads only the keys of levels $\ell \ge 1$ and runs entirely on the GPU. A gather kernel then copies the exact
tokens in the candidate list of level 0, the children of the $k$ selected level-1 nodes, from host memory over PCIe in
16-token blocks (\ding{175}). Each step gathers at most $Ck = 4{,}096$ exact tokens per KV head and layer, at most
576~MiB over the 36 layers, independent of the context length.

After space is reserved for the generated tokens, the remaining GPU memory caches a suffix of the input. The gather kernel reads the cached blocks from the GPU cache and the
others from host memory.

On the RULER samples of Appendix~D.1, decoding at 1M and 2M takes 8.35 and 10.7~ms per token with the cache,
compared with 19.8 and 20.6~ms without the cache.

\section{Evaluation}
\label{sec:eval}

\subsection{Experimental Setup}
\label{sec:setup}

\paragraph{Model and training.}
We apply \recsa to Qwen3-4B~\citep{qwen2025qwen3}, whose training length is 32K, and train it with the two stages of
\S\ref{sec:training} on 0.62B tokens in total. The training data include synthetic retrieval tasks, document question answering, books, code and instructions
whose responses were generated by the original model (Appendix~E). Unless otherwise stated, we evaluate with the default $k = 256$.

\paragraph{Baselines.}
We train three backbones, Dense, MoBA and DSA, from Qwen3-4B with the settings, sequences, order and number of
steps of Stage~2; for DSA, the indexer is first warmed up as in \citet{deepseek2025v32}. We evaluate each backbone with YaRN
and with DCA and name each model by its attention and its context extension method, such as Dense~+~YaRN.
Qwen3-4B~+~YaRN denotes the original model with YaRN. The training-free InfLLM and InfiniteHiP run on the original
model with their official implementations. At each length, the token budget of every sparse baseline, counted as in
\S\ref{sec:background}, is matched to that of \recsa for the query at the last position (Table~\ref{tab:ruler}).
Appendix~E.4 gives the settings at each length.

\begin{table}[!t]
\caption{RULER score (average over the 13 tasks) by context length. Token budget: entries the query at the last position attends to (for
\recsa, exact tokens plus summaries), matched across the sparse methods. Best results are in bold.}
\label{tab:ruler}
\vskip 0.15in
\centering
\small
{\setlength{\tabcolsep}{2.5pt}%
\begin{tabular}{lrrrrrr}
\toprule
Method & 32K & 131K & 262K & 524K & 1M & 2M \\
\textit{Token budget} & 5,888 & 8,192 & 8,704 & 9,728 & 11,575 & 11,990 \\
\textit{\% of context} & 18.0 & 6.25 & 3.32 & 1.86 & 1.16 & 0.60 \\
\midrule
\multicolumn{7}{l}{\textit{Training-free}} \\
\quad InfLLM & 45.0 & 29.3 & 29.2 & 27.4 & 25.7 & 24.2 \\
\quad InfiniteHiP & 79.6 & 63.7 & 57.7 & 49.4 & 36.6 & 34.0 \\
\midrule
\multicolumn{7}{l}{\textit{Trained at 32K}} \\
\quad Dense + YaRN & 94.5 & 80.0 & 48.3 & 3.6 & 1.6 & 1.1 \\
\quad Dense + DCA & 94.6 & 78.5 & 68.4 & 58.0 & 40.0 & 27.6 \\
\quad MoBA + YaRN & 81.6 & 31.9 & 5.1 & 0.9 & 1.0 & 0.6 \\
\quad MoBA + DCA & 81.8 & 37.9 & 23.8 & 18.1 & 12.5 & 7.2 \\
\quad DSA + YaRN & \textbf{95.3} & 80.5 & 57.1 & 18.8 & 6.0 & 1.4 \\
\quad DSA + DCA & 95.2 & 77.5 & 65.7 & 55.2 & 36.8 & 22.6 \\
\addlinespace[2pt]
RecSA (ours) & 94.7 & \textbf{83.3} & \textbf{75.9} & \textbf{69.7} & \textbf{60.5} & \textbf{49.3} \\
\bottomrule
\end{tabular}
}
\vskip -0.1in
\end{table}

\paragraph{Benchmarks.}
RULER~\citep{hsieh2024ruler} has 13 synthetic tasks (Appendix~B). We report its official score, averaged over all tasks, with greedy
decoding on 56 samples per task from 32K to 262K, 20 at 524K and 16 at 1M and above. All methods in
Table~\ref{tab:ruler} use the same samples. Perplexity at context length $L$ is computed on the last $L$ of the first $10^6$ tokens of each of nine
PG-19 books~\citep{rae2020pg19} (Appendix~C). For long-document tasks, we select LongBench~\citep{bai2024longbench} and En.MC from
InfiniteBench~\citep{zhang2024infinitebench}, and for short-context tasks, MMLU~\citep{hendrycks2021mmlu}, ARC-C~\citep{clark2018arc}, GSM8K~\citep{cobbe2021gsm8k},
\mbox{IFEval}~\citep{zhou2023ifeval}, HumanEval~\citep{chen2021humaneval} and MBPP~\citep{austin2021mbpp}. Context lengths
from 32K to 524K denote $2^{15}$ to $2^{19}$ tokens, and 1M and 2M denote $10^6$ and $2 \times 10^6$ tokens.

\begin{table*}[!t]
\caption{RULER score per task at 131K and 1M. Retrieval: needle-in-a-haystack tasks with a single needle (S1--S3), multiple keys (MK1--MK3), multiple values (MV)
or multiple queries (MQ). VT: variable tracking. CWE: common words extraction. FWE: frequent words extraction. QA1:
SQuAD~\citep{rajpurkar2016squad}. QA2: HotpotQA~\citep{yang2018hotpotqa}. Best results in each column at each length
are in bold. Appendix~B gives the scores at all lengths.}
\label{tab:ruler-tasks-main}
\vskip 0.15in
\centering
\footnotesize
{\setlength{\tabcolsep}{2pt}%
\begin{tabular*}{\textwidth}{@{\extracolsep{\fill}}lrrrrrrrrrrrrrr}
\toprule
 & \multicolumn{8}{c}{Retrieval} & \multicolumn{1}{c}{Tracing} & \multicolumn{2}{c}{Aggregation} & \multicolumn{2}{c}{QA} & \\
\cmidrule(lr){2-9}\cmidrule(lr){10-10}\cmidrule(lr){11-12}\cmidrule(lr){13-14}
Method & S1 & S2 & S3 & MK1 & MK2 & MK3 & MV & MQ & VT & CWE & FWE & QA1 & QA2 & Avg \\
\midrule
\multicolumn{15}{l}{\textit{131K}} \\
InfLLM & \textbf{100.0} & 7.1 & 0.0 & 5.4 & 57.1 & 0.0 & 7.6 & 3.6 & 60.0 & 36.4 & 86.3 & 8.9 & 8.9 & 29.3 \\
InfiniteHiP & \textbf{100.0} & 91.1 & 91.1 & 73.2 & 32.1 & 0.0 & 76.3 & 86.6 & 57.1 & 55.9 & 86.3 & 55.4 & 23.2 & 63.7 \\
\addlinespace[2pt]
Dense + YaRN & \textbf{100.0} & \textbf{100.0} & 98.2 & 94.6 & 64.3 & \textbf{41.1} & 91.5 & 94.6 & 95.7 & 70.7 & 85.7 & 58.9 & 44.6 & 80.0 \\
Dense + DCA & \textbf{100.0} & 98.2 & \textbf{100.0} & 78.6 & 73.2 & \textbf{41.1} & 81.7 & 94.6 & 90.4 & 79.3 & \textbf{90.5} & 60.7 & 32.1 & 78.5 \\
MoBA + YaRN & 87.5 & 10.7 & 25.0 & 19.6 & 14.3 & 1.8 & 18.8 & 8.9 & 81.8 & 41.8 & 58.3 & 30.4 & 16.1 & 31.9 \\
MoBA + DCA & 98.2 & 14.3 & 17.9 & 19.6 & 23.2 & 8.9 & 17.9 & 12.1 & 81.8 & 48.2 & 88.7 & 37.5 & 25.0 & 37.9 \\
DSA + YaRN & \textbf{100.0} & \textbf{100.0} & \textbf{100.0} & 94.6 & 66.1 & 26.8 & 88.8 & 96.4 & 98.2 & 80.5 & 84.5 & \textbf{67.9} & 42.9 & 80.5 \\
DSA + DCA & \textbf{100.0} & 98.2 & \textbf{100.0} & 78.6 & 69.6 & 37.5 & 79.0 & 92.9 & 89.3 & 81.8 & 86.3 & 62.5 & 32.1 & 77.5 \\
\addlinespace[2pt]
RecSA (ours) & \textbf{100.0} & \textbf{100.0} & \textbf{100.0} & \textbf{100.0} & \textbf{76.8} & 37.5 & \textbf{92.0} & \textbf{99.6} & \textbf{99.6} & \textbf{84.8} & 88.7 & 57.1 & \textbf{46.4} & \textbf{83.3} \\
\midrule
\multicolumn{15}{l}{\textit{1M}} \\
InfLLM & \textbf{100.0} & 6.2 & 0.0 & 0.0 & \textbf{56.2} & 0.0 & 1.6 & 3.1 & 53.8 & 7.5 & 81.2 & 18.8 & 6.2 & 25.7 \\
InfiniteHiP & 87.5 & 62.5 & 43.8 & 37.5 & 6.2 & 0.0 & 42.2 & 37.5 & 52.5 & 1.9 & \textbf{85.4} & 18.8 & 0.0 & 36.6 \\
\addlinespace[2pt]
Dense + YaRN & 0.0 & 0.0 & 0.0 & 0.0 & 0.0 & 0.0 & 0.0 & 0.0 & 0.0 & 0.0 & 20.8 & 0.0 & 0.0 & 1.6 \\
Dense + DCA & \textbf{100.0} & 56.2 & 75.0 & 56.2 & 6.2 & 0.0 & 29.7 & 26.6 & 82.5 & 2.5 & 79.2 & 6.2 & 0.0 & 40.0 \\
MoBA + YaRN & 0.0 & 0.0 & 0.0 & 0.0 & 0.0 & 0.0 & 0.0 & 0.0 & 0.0 & 0.0 & 12.5 & 0.0 & 0.0 & 1.0 \\
MoBA + DCA & 75.0 & 0.0 & 0.0 & 0.0 & 0.0 & 0.0 & 0.0 & 0.0 & 15.0 & 0.0 & 60.4 & 6.2 & 6.2 & 12.5 \\
DSA + YaRN & 43.8 & 0.0 & 0.0 & 0.0 & 0.0 & 0.0 & 0.0 & 0.0 & 1.2 & 0.0 & 27.1 & 6.2 & 0.0 & 6.0 \\
DSA + DCA & \textbf{100.0} & 31.2 & 62.5 & 43.8 & 6.2 & 0.0 & 29.7 & 20.3 & 71.2 & 21.9 & 72.9 & 6.2 & \textbf{12.5} & 36.8 \\
\addlinespace[2pt]
RecSA (ours) & \textbf{100.0} & \textbf{87.5} & \textbf{100.0} & \textbf{81.2} & 25.0 & \textbf{6.2} & \textbf{54.7} & \textbf{90.6} & \textbf{95.0} & \textbf{23.1} & \textbf{85.4} & \textbf{31.2} & 6.2 & \textbf{60.5} \\
\bottomrule
\end{tabular*}
}
\vskip -0.1in
\end{table*}

\paragraph{Efficiency.}
We measure latency on one H200 GPU with batch size~1. In the same framework as \recsa, Dense uses FlashAttention-2~\citep{dao2024flashattention2} for
prefill and Flash-Decoding~\citep{dao2023flashdecoding} for decoding. At 1M and 2M, host memory holds the exact tokens of \recsa (\S\ref{sec:offload}) and, in decoding, the part of the Dense KV
cache beyond GPU memory. Appendix~D.1 describes how we measure latency.

\subsection{Length Extrapolation}
\label{sec:extrapolation}

\begin{table}[!t]
\caption{Perplexity on nine PG-19 books~\citep{rae2020pg19} by context length. Sparse methods use the token budgets of
Table~\ref{tab:ruler}. InfLLM and InfiniteHiP are not listed because their key selection during prefill is not
causal. Best results are in bold.}
\label{tab:ppl}
\vskip 0.15in
\centering
\small
{\setlength{\tabcolsep}{4pt}%
\begin{tabular}{lrrrrr}
\toprule
Method & 32K & 131K & 262K & 524K & 1M \\
\midrule
\multicolumn{6}{l}{\textit{Trained at 32K}} \\
\quad Dense + YaRN & 11.17 & 11.81 & 16.96 & 60.94 & 211.33 \\
\quad Dense + DCA & 11.17 & 11.27 & 12.20 & 13.39 & 14.57 \\
\quad MoBA + YaRN & 12.11 & 13.51 & 20.43 & 122.30 & 414.69 \\
\quad MoBA + DCA & 12.11 & 12.87 & 14.85 & 17.12 & 19.18 \\
\quad DSA + YaRN & \textbf{11.15} & 11.81 & 15.23 & 25.88 & 54.42 \\
\quad DSA + DCA & \textbf{11.15} & 11.32 & 12.35 & 13.67 & 15.01 \\
\addlinespace[2pt]
RecSA (ours) & 11.24 & \textbf{10.87} & \textbf{11.00} & \textbf{11.08} & \textbf{11.12} \\
\bottomrule
\end{tabular}}

\vskip -0.1in
\end{table}

Table~\ref{tab:ruler} shows that \recsa achieves the highest RULER score at every length from 131K to 2M. It outperforms
the strongest baseline by 2.8 points at 131K (DSA~+~YaRN) and by 11.7 at 524K and 20.5 at 1M (Dense~+~DCA). At 2M,
\recsa scores 49.3, compared with 34.0 for InfiniteHiP~\citep{lee2025infinitehip}, the strongest baseline at this length. At 32K, \recsa matches
Dense~+~YaRN (94.7 vs.\ 94.5).

Beyond 131K, each backbone scores higher with DCA than with YaRN. For example, Dense~+~DCA scores 40.0 at 1M and
Dense~+~YaRN 1.6. Like \recsa, DSA~+~DCA uses a matched token budget and keeps every relative distance within the range seen during
training, but at 1M it scores 36.8, compared with 60.5 for \recsa. Keeping relative distances within this range thus
does not by itself explain the higher score of \recsa.

On individual tasks (Table~\ref{tab:ruler-tasks-main} and Appendix~B), \recsa achieves the highest score
on MK1, MQ, MV, S2 and VT at every length from 131K to 2M. For example, at 1M, \recsa scores 90.6 on
multi-query retrieval (MQ) and 95.0 on variable tracking (VT); the strongest baselines on these two tasks score 37.5 (InfiniteHiP) and 82.5 (Dense~+~DCA).

From 131K to 1M, the perplexity of \recsa stays between 10.87 and 11.12 (Table~\ref{tab:ppl}). At 1M, the perplexity ranges from
54.42 to 414.69 for the methods with YaRN and from 14.57 to 19.18 for those with DCA.

\begin{figure*}[!t]
\centering
\includegraphics[width=\textwidth]{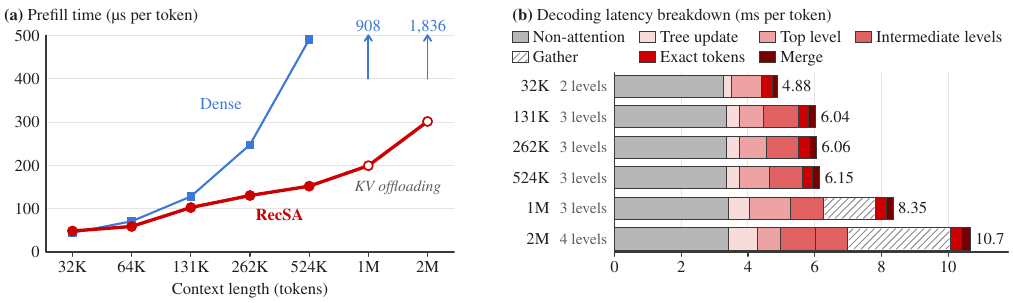}\par
\caption{Efficiency on one H200 GPU with batch size~1. (a)~Prefill time per token of \recsa and of Dense with
FlashAttention in the same framework; hollow markers indicate \recsa with offloading (\S\ref{sec:offload}), and arrows
mark values above the axis range. (b)~Time breakdown of one decoding step of \recsa. Non-attention: the remaining step time, including the
projections, the MLPs and the idle time between kernels. Gather: gathering the selected exact tokens from host
memory and from the GPU cache (\S\ref{sec:offload}).}
\label{fig:eff}
\end{figure*}

\begin{figure*}[!t]
\centering
\includegraphics[width=\textwidth]{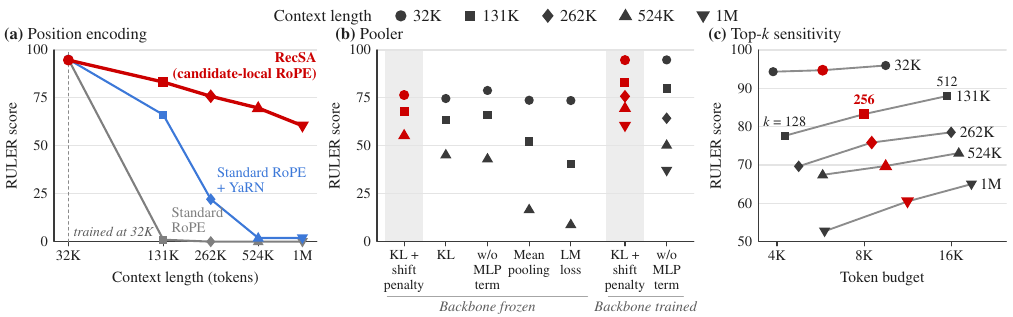}\par
\caption{RULER score in the ablation experiments. (a)~Position encoding. Standard RoPE applies RoPE at absolute token
positions and is evaluated with and without YaRN beyond 32K. (b)~Pooler. KL and the shift penalty are the losses of Eqs.~\eqref{eq:kl} and~\eqref{eq:shift}, and LM loss trains the MLP term with the language modeling loss. Backbone frozen: each pooler is trained on the original model and evaluated with it at 32K, 131K and 524K. Backbone trained: for KL + shift penalty and w/o MLP term,
the backbone is then trained as in Stage~2.
KL + shift penalty is the pooler of \recsa. (c)~\recsa evaluated with $k = 128$, 256 and 512.
Appendix~F gives all scores.}
\label{fig:ablation}
\end{figure*}

\subsection{Short-Context and Long-Document Tasks}
\label{sec:general}

\begin{table}[!t]
\caption{Scores on short-context benchmarks and long-document tasks. LongBench~\citep{bai2024longbench}: average score
over 12 English tasks and 2 code tasks. En.MC: multiple-choice questions about full-length novels from
InfiniteBench~\citep{zhang2024infinitebench}, with inputs of 70K--746K tokens; Qwen3-4B uses YaRN for context extension
on En.MC. Best results in each row are in bold.}
\label{tab:general}
\vskip 0.15in
\centering
\small
\begin{tabular}{lrr}
\toprule
Benchmark & Qwen3-4B & RecSA \\
\midrule
MMLU (5-shot) & 70.1 & \textbf{70.5} \\
ARC-C (0-shot) & 87.2 & \textbf{88.1} \\
GSM8K (0-shot, CoT) & \textbf{91.5} & 90.7 \\
IFEval (prompt, strict) & \textbf{79.1} & 78.6 \\
HumanEval (0-shot, pass@1) & \textbf{82.3} & 81.7 \\
MBPP (0-shot, pass@1) & \textbf{63.6} & 61.8 \\
\midrule
LongBench & 49.2 & \textbf{51.8} \\
InfiniteBench En.MC & 34.5 & \textbf{66.8} \\
\bottomrule
\end{tabular}

\vskip -0.1in
\end{table}

On the six short-context benchmarks (Table~\ref{tab:general}), \recsa performs
comparably to Qwen3-4B; the largest difference is 1.8 points, on MBPP (61.8 vs.\ 63.6). On LongBench, \recsa scores
51.8, compared with 49.2 for the original Qwen3-4B and 50.9 for the Dense backbone. On En.MC, \recsa scores 66.8 and Qwen3-4B~+~YaRN 34.5;
InfiniteHiP, Dense~+~DCA and DSA~+~DCA score 66.4, 62.4 and 62.0, respectively. Appendix~C gives the scores of Dense on
all benchmarks and the En.MC scores by input length.

\subsection{Efficiency}
\label{sec:efficiency}

Prefill of 32K tokens takes 1.59~s with \recsa and 1.46~s with Dense, and from 64K to 2M,
\recsa is faster (Figure~\ref{fig:eff}a). At 524K, prefill takes 80~s with \recsa and 257~s with Dense, a $3.2\times$
speedup.

\recsa is faster than Dense in decoding at every measured length, including 32K
(Figure~\ref{fig:teaser}b).
From 131K to 524K, \recsa takes about 6.1~ms per token, while the latency of Dense grows from 8.6~ms to 23.4~ms,
$3.8\times$ that of \recsa at 524K. At 1M and 2M, \recsa takes 8.35 and 10.7~ms per token, compared with 177~ms and 2.9~s for
Dense. Copying the part of the Dense KV cache in host memory to the GPU accounts for 77\% and 96\% of the Dense latency,
respectively.

The decoding step time of \recsa (Figure~\ref{fig:eff}b) stays nearly constant from
131K to 524K, where the tree has three levels and each level scores at most $Ck = 4{,}096$ candidates
(\S\ref{sec:tree}). At these lengths, the non-attention part takes 3.35~ms, more than half of the step. At 1M and 2M,
the selected exact tokens are gathered from host memory and from the GPU cache (Gather), and at 2M, the tree has a fourth level.
A two-level variant scores every level-1 node at each step,
8{,}192 nodes at 131K and 125{,}000 at 2M, and \recsa is $1.1\times$ faster than this variant at 131K and $1.9\times$
faster at 2M (Table~D.3).

\subsection{Ablation Studies}
\label{sec:ablation}

Unless otherwise stated, each variant below is trained with the same data, order and steps as \recsa.

\paragraph{Summaries improve the average score at long contexts.}
A model trained and evaluated without summaries, whose softmax covers only the exact tokens, averages 70.4 over the
four lengths from 131K to 1M, compared with 72.3 for \recsa (Appendix~F).

\paragraph{Candidate-local RoPE improves length extrapolation.}
The standard RoPE variant is trained with both stages of \S\ref{sec:training}, the pooler included, but applies RoPE at
absolute token positions. At 32K, it scores 95.2 and \recsa 94.7 (Figure~\ref{fig:ablation}a). Without YaRN, standard RoPE
scores 1.0 at 131K, where relative distances exceed those seen during training. With YaRN, it scores 66.2 at 131K and
1.9 at 524K and 1M, while \recsa scores 83.3 at 131K and 60.5 at 1M.

\newpage

\paragraph{Distillation, the shift penalty and the MLP term improve the score at long contexts.}
With the backbone frozen (Figure~\ref{fig:ablation}b), the five poolers score between 73.6 and 78.9 at 32K but between
8.9 and 55.5 at 524K, where the pooler of \recsa (KL + shift penalty) scores highest. Training the MLP term with the language
modeling loss instead of distillation gives the lowest scores at 131K and 524K, below mean pooling, and training with KL alone,
without the shift penalty, lowers the score at 524K from 55.5 to 45.3. With the backbone trained, the decrease from removing the MLP term
grows with length: 3.3 points at 131K, 11.5 at 262K, 19.3 at 524K and 23.4 at 1M. Appendix~A.4 examines how the MLP term changes
selection.

\paragraph{Increasing $k$ to 512 improves the score at every length.}
Because $k$ sets the token budget (\S\ref{sec:tree}), we evaluate the trained \recsa model with $k = 128$ and $k = 512$
(Figure~\ref{fig:ablation}c). At every length from 32K to 1M, $k = 512$ scores higher than $k = 256$, with
$1.7\times$--$1.9\times$ the token budget, and $k = 128$ scores lower at 131K, 262K and 1M. With $k = 512$,
relative distances in a candidate list reach 8{,}191, compared with at most 4{,}095 in training with $k = 256$.

\section{Related Work}
\label{sec:related}

\paragraph{Hierarchical selection and summaries.}
HiP~\citep{lee2025hip} is training-free and selects level by level in a binary tree, scoring each block by one
representative key. PISA~\citep{tang2026pisa} builds a pyramid of mean-pooled block keys, scores each candidate by the
LogSumExp of the scores of its children and trains models with this selection. Both discard the unselected blocks.
Other methods combine attention over selected tokens with coarse representations of the context. NSA attends to the compressed blocks of the whole
context and to the selected blocks in separate branches. Multipole Attention~\citep{hooper2025multipole} is
training-free: in decoding, it clusters the keys with k-means, attends exactly to the keys of high-scoring clusters,
approximates the other clusters by their centroids and normalizes both parts together. \recsa also normalizes summaries
and exact tokens together, and they partition the prefix; the keys of its summaries come from a trained pooler,
and each level scores a bounded number of candidates.

\paragraph{Sparse attention beyond the training length.}
Landmark Attention~\citep{mohtashami2023landmark}, HSA~\citep{hu2025hsa,leng2026hsalength} and
HiLS~\citep{hu2026hils} learn a representation for each block and train the block scores end to end with the
language modeling loss. HSA pretrains models of about 240M parameters from scratch at 4K, fine-tunes them on RULER
tasks and evaluates them up to 32M. HiLS pretrains models from scratch at 8K and evaluates them up to 4M; it also
converts the pretrained Olmo3-7B to HiLS by continued pretraining at 8K, with RULER examples mixed into the training
data, and evaluates it on RULER up to 1M.

\vspace{-0.3em}
\paragraph{Distilled selectors.}
SeerAttention~\citep{gao2024seerattention} fits a block-level gate to the max-pooled attention map with a KL divergence and uses it in prefill;
SeerAttention-R~\citep{gao2025seerattentionr} applies such a gate in decoding. The warm-up stage of
DSA~\citep{deepseek2025v32} trains only the indexer, fitting it to dense attention with a KL divergence,
before the whole model is trained. These methods supervise the scores of one level. \recsa
supervises the scores of every level with the dense attention mass on the subtree of each candidate, through one pooler
shared across levels, and the keys that the pooler computes serve both for selection and as the keys of summaries.

\vspace{-0.3em}
\paragraph{Position encoding.}
InfLLM~\citep{xiao2024infllm} assigns one position to all tokens beyond the local window, InfiniteHiP gives the
selected keys consecutive positions and Landmark Attention maps the retrieved blocks into a reserved range of
positions. HiLS keeps RoPE only in the dimensions whose rotation period does not exceed the pretraining length; PISA
applies RoPE to the high-frequency half of the dimensions of each head. Context extension methods, such as YaRN, DCA,
ReRoPE, PI~\citep{chen2023pi}, Long\-RoPE~\citep{ding2024longrope} and SelfExtend~\citep{jin2024selfextend},
change the rotation angles or the assignment of positions. \recsa assigns each candidate its index in the candidate list
of its level, and each level has at most $Ck$ candidates, so every relative distance stays below $Ck$ at any context
length.

\vspace{-0.3em}
\paragraph{Training-free sparse attention and KV offloading.}
Training-free sparse methods accelerate prefill~\citep{jiang2024minference,lai2025flexprefill} or decoding.
Quest~\citep{tang2024quest} estimates, at each decoding step, an upper bound on the query--key scores in every page of
the KV cache, and the first pruning stage of InfiniteHiP scans all keys; in \recsa, the selection cost per query is
$O(\log L)$. InfLLM and \mbox{InfiniteHiP}, like \recsa (\S\ref{sec:offload}), keep most of the KV cache in host memory and
cache part of it on the GPU.

\section{Conclusion}
\label{sec:conclusion}

We present \recsa, which builds a tree over the KV cache and selects level by level; the summaries of unselected
candidates share one softmax with the exact tokens. Its selection cost and the number of entries read per query are
$O(\log L)$, and candidate-local RoPE keeps every relative distance within the range seen during training, so \recsa has
no inherent limit on context length. We initialize \recsa from Qwen3-4B, distill its pooler from dense attention and
train it only at the original model's 32K training length. On RULER, \recsa outperforms all baselines
from 131K to 2M. With our kernels, \recsa is faster than dense attention in end-to-end prefill and decoding from 64K to
2M.

\clearpage %
\bibliography{references}
\bibliographystyle{mlsys2025}

\section*{AI Usage Disclosure}
We used AI coding assistants for four tasks. Code generation: they wrote parts of the implementation to the authors'
specifications, including the training and evaluation scripts. Experiment execution: they launched and monitored the
training and evaluation jobs that the authors specified and approved, and collected the results. Literature search:
they helped find related work and check citation metadata against the original sources. Writing: they polished the
grammar and phrasing of the authors' text. Separately, some of the training data consists of responses generated by
the original Qwen3-4B (Appendix~E). The research ideas, the method, the experimental design, the analysis and
the conclusions are the authors' own. The authors reviewed all AI-assisted work and take full responsibility for the
content of this paper.

\clearpage
\makeatletter
\let\@addtocurcol\appx@addtocurcol
\setlength{\@dblfptop}{0pt}
\setlength{\@fptop}{0pt}
\makeatother
\counterwithin{table}{section}
\counterwithin{figure}{section}
\hyphenation{InfiniteHiP}
\raggedbottom

\appendix
\section{Tree and Pooler}
\label{app:pooler}

The pooler of each attention layer has the parameters $\mathbf{w} \in \mathbb{R}^{128}$, $\tau$, $g$,
$W_1 \in \mathbb{R}^{64 \times 256}$, $\mathbf{b}_1 \in \mathbb{R}^{64}$ and $W_2 \in \mathbb{R}^{128 \times 64}$
(Eq.~(5)), and clamps $\sigma$ and $\rho$ to a minimum of $10^{-6}$.

In Stage~1, the candidate lists in the KL (Eq.~(6)) include the current node, whose key is pooled
from all of its children (Eq.~(1)).

\subsection{Nodes with Fewer than $C$ Children}
\label{app:pooler-partial}
The last node of each level $\ell \ge 1$ has $n < C$ children until it is full. For such a node, Eqs.~(1) and~(5) use only these $n$ children: the value is the mean of their values, $\boldsymbol{\mu}$ and $\sigma$ are computed over them and the softmax that gives $\alpha_c$ runs over them.

\subsection{Tree Update in Decoding}
\label{app:pooler-decode}
In decoding, Recursive Sparse Attention (\recsa) recomputes the key and value of the node that contains the new token at each
level, from level~1 up to the top level (Section~4.2). When the position $t$ of the new token is a
multiple of $C^{\ell}$, \recsa adds a node with one child at level $\ell$. When the top level exceeds $Ck$ nodes, \recsa builds a new top level from it by Eq.~(1), and the existing levels do not change.

\subsection{Initialization}
\label{app:pooler-init}
The first part of Stage~1 initializes the pooler of every layer as follows, with separate random draws for each layer:
$\mathbf{w}$ is a unit vector in a random direction, $\tau = 0.05$, $g = 1$, the elements of $W_1$ are drawn
independently and uniformly from $[-1/16, 1/16]$, $\mathbf{b}_1 = \mathbf{0}$ and $W_2 = 0$. The initial key is then
$\mathbf{p}$, and the small $\tau$ keeps every $\alpha_c$ close to $1/C$, so this key is close to the mean of the child
keys. The first part leaves $W_1$, $\mathbf{b}_1$ and $W_2$ at these values, and the second part trains them from
there.

\subsection{Analysis of the MLP Term}
\label{app:pooler-mlp}

We use the notation of Eq.~(5) and write $\mathbf{q}$ for a query vector. With $g = 1$, the score $\mathbf{q}^\top \mathbf{K}$ without the MLP
term equals, up to a constant, the first-order Taylor expansion of $\log \sum_c \exp(\mathbf{q}^\top \mathbf{x}_c)$ at
$\mathbf{q} = \tau \mathbf{w} / \sigma$~\citep{hu2026hils}, because $\mathbf{p}$ is the gradient of this LogSumExp at that point. No single key can match
this LogSumExp for every query, because $\mathbf{q}^\top \mathbf{K}$ is linear in $\mathbf{q}$ and the LogSumExp is
not. The MLP term can move the key away from $\mathbf{p}$, and we measure what it contributes.

\begin{table}[t]
\caption{The MLP term on needle nodes, the nodes that contain evidence for the answer. 131K, 28 RULER samples, queries
of the first four answer tokens; w/o MLP term: $W_2 = 0$. $\alpha$: the sum of the weights of the level-1 needle nodes
among the 16 children of a level-2 needle node (0.14 if all children had equal weight). LSE: LogSumExp of the token
scores below a node; the row LSE $-$ node score is for level-2 needle nodes; scores are centered over the nodes of each
query. Recall: fraction of queries for which a level-2 needle node is
among the top 1/8 of the level-2 nodes (0.635 when nodes are ranked by LSE).}
\label{tab:pooler-mlp}
\vskip 0.15in
\centering
\small
\setlength{\tabcolsep}{4pt}
\begin{tabular}{lrr}
\toprule
 & w/o MLP term & \recsa \\
\midrule
$\alpha$ on level-1 needle nodes & 0.23 & 0.52 \\
LSE $-$ node score & 1.21 & 0.34 \\
Recall & 0.468 & 0.625 \\
\bottomrule
\end{tabular}
\vskip -0.1in
\end{table}

\paragraph{Setup.} We used 28 RULER~\citep{hsieh2024ruler} samples at 131K, four from each of S2, S3, MK1, MK2, MV, MQ and
VT (Table~2). For each sample, we ran \recsa with the correct answer appended and kept the keys of every
layer and the queries of the first four answer tokens. From these keys we recomputed levels 1 (16 tokens) and 2
(256 tokens) with and without the MLP term ($W_2 = 0$, all other parameters unchanged). We compared the score of each node
with the LogSumExp (LSE) of the query's scores over all tokens below it. Selection does not change when the scores of
all nodes are shifted by the same amount, so we centered the scores over the nodes of each query. We call a node that contains
evidence for the answer a \emph{needle node}. We measure recall as the fraction of queries for which a needle node is
among the top 1/8 of the nodes at its level.

\paragraph{Results.} The MLP term does not correct the score of the node whose key it enters. Its contribution to that
score, $\sigma \mathbf{q}^\top \mathbf{m}$, has a correlation below 0.1 in magnitude with the gap between
the LSE and the mean score of the tokens below the node at both levels, and level-1 recall is unchanged (0.747 vs.\ 0.752). Its effect appears at the next level. The MLP term
of a level-1 key enters the level-2 weights $\alpha_c$ through $\mathbf{w}^\top (\mathbf{x}_c - \boldsymbol{\mu})$ and acts there as
a query-independent bias: the logits of the level-1 needle nodes rise by 1.45, the sum of their
weights increases from 0.23 to 0.52 (Table~\ref{tab:pooler-mlp}) and the level-2 key moves closer to them. Of the 0.878 by which the MLP term raises the score of a needle node at
level 2, the change in $g\,(\mathbf{p} - \boldsymbol{\mu})$ contributes 1.025 and the terms $\sigma \mathbf{m}$ of both levels $-0.147$. With the MLP term at level 2 only, level-2
recall is 0.476, compared with 0.468 without the term; with the term at level 1 only, level-2 recall is 0.628. The keys
of the original Qwen3-4B~\citep{qwen2025qwen3}, on which the
pooler is trained, give nearly the same numbers on 14 of these samples (two per task): the gap between the LSE and the node score falls from 1.23 to 0.37,
and recall rises from 0.469 to 0.628.

\section{RULER Results per Task}
\label{app:ruler}
\paragraph{Setup.} We generate the samples with our reimplementation of the RULER generators~\citep{hsieh2024ruler},
which differs from the official code in four ways. The essay haystack consists of web documents from
DCLM-Baseline~\citep{li2024dclm} instead of the essays of Paul Graham. The uncommon words of CWE and the words of FWE
are random strings of three consonant--vowel syllables, where the official code uses English words for CWE and random
strings of six letters for FWE. The distractor keys of MK2 carry a numeric suffix, so that they stay unique at long
contexts. QA1 draws its questions from SQuAD v1.1 instead of SQuAD 2.0. From 524K on, QA1 needs more distractor
paragraphs than the development set of SQuAD v1.1 has, so the distractor paragraphs repeat. Each input is one user
turn in the chat template of Qwen3-4B in non-thinking mode, and the assistant turn starts with the answer prefix of the
task; with the template, an input has 95\% to 100\% of $L$ tokens. Each task keeps the official limit on generated
tokens: 128 for S1 to MQ, 30 for VT, 120 for CWE, 50 for FWE and 32 for QA1 and QA2. The score is the string match of
the official code: the fraction of gold answers that appear in the output or, for QA1 and QA2, whether any gold answer
appears.

\paragraph{Results.} Tables~\ref{tab:ruler-tasks} and~\ref{tab:ruler-tasks-2} list the score of each task, with the abbreviations of
Table~2, for the methods of Table~1 at every length. They also list the original
Qwen3-4B with YaRN~\citep{peng2024yarn} up to 524K, the original Qwen3-4B with DCA~\citep{an2024dca} up to 1M and
the variants of Table~\ref{tab:ablation} with a trained backbone up to 1M.

\begin{table*}[p]
\caption{RULER score per task, 32K--262K. The best Avg among the methods of Table~1 is in bold.}
\label{tab:ruler-tasks}
\vskip 0.15in
\centering
\footnotesize
{\setlength{\tabcolsep}{2pt}%
\begin{tabular*}{\textwidth}{@{\extracolsep{\fill}}lrrrrrrrrrrrrrr}
\toprule
\rlap{32K}\phantom{standard RoPE + YaRN} & S1 & S2 & S3 & MK1 & MK2 & MK3 & MV & MQ & VT & CWE & FWE & QA1 & QA2 & Avg \\
\midrule
InfLLM & 100.0 & 17.9 & 32.1 & 32.1 & 67.9 & 10.7 & 29.0 & 36.6 & 57.1 & 56.1 & 81.0 & 39.3 & 25.0 & 45.0 \\
InfiniteHiP & 100.0 & 98.2 & 100.0 & 96.4 & 58.9 & 19.6 & 95.1 & 99.6 & 53.2 & 76.2 & 82.7 & 87.5 & 67.9 & 79.6 \\
Dense + YaRN & 100.0 & 98.2 & 100.0 & 100.0 & 94.6 & 100.0 & 99.6 & 100.0 & 99.3 & 93.2 & 89.9 & 87.5 & 66.1 & 94.5 \\
Dense + DCA & 100.0 & 98.2 & 100.0 & 100.0 & 96.4 & 100.0 & 99.6 & 100.0 & 99.3 & 93.4 & 89.9 & 87.5 & 66.1 & 94.6 \\
MoBA + YaRN & 100.0 & 71.4 & 96.4 & 76.8 & 76.8 & 89.3 & 67.0 & 72.3 & 94.6 & 86.6 & 90.5 & 82.1 & 57.1 & 81.6 \\
MoBA + DCA & 100.0 & 71.4 & 91.1 & 78.6 & 80.4 & 87.5 & 64.7 & 77.7 & 95.4 & 85.7 & 91.1 & 82.1 & 57.1 & 81.8 \\
DSA + YaRN & 100.0 & 100.0 & 100.0 & 100.0 & 94.6 & 100.0 & 100.0 & 100.0 & 98.9 & 94.6 & 91.7 & 91.1 & 67.9 & \textbf{95.3} \\
DSA + DCA & 100.0 & 98.2 & 100.0 & 100.0 & 94.6 & 100.0 & 100.0 & 100.0 & 99.3 & 95.2 & 91.1 & 91.1 & 67.9 & 95.2 \\
\addlinespace[2pt]
RecSA (ours) & 100.0 & 100.0 & 100.0 & 98.2 & 96.4 & 96.4 & 98.2 & 100.0 & 100.0 & 95.2 & 89.9 & 92.9 & 64.3 & 94.7 \\
\midrule
Qwen3-4B + YaRN & 100.0 & 96.4 & 100.0 & 98.2 & 89.3 & 76.8 & 98.7 & 99.1 & 54.3 & 73.2 & 83.3 & 83.9 & 60.7 & 85.7 \\
Qwen3-4B + DCA & 100.0 & 98.2 & 100.0 & 100.0 & 89.3 & 78.6 & 98.7 & 99.1 & 53.6 & 72.7 & 82.7 & 83.9 & 62.5 & 86.1 \\
\addlinespace[2pt]
w/o summaries & 100.0 & 100.0 & 100.0 & 100.0 & 94.6 & 98.2 & 98.2 & 100.0 & 99.6 & 93.6 & 86.9 & 91.1 & 66.1 & 94.5 \\
w/o MLP term & 100.0 & 100.0 & 100.0 & 98.2 & 98.2 & 98.2 & 97.8 & 100.0 & 100.0 & 95.9 & 89.3 & 87.5 & 67.9 & 94.8 \\
standard RoPE + YaRN & 100.0 & 98.2 & 100.0 & 100.0 & 94.6 & 100.0 & 98.7 & 100.0 & 100.0 & 93.8 & 91.1 & 92.9 & 67.9 & 95.2 \\
standard RoPE & 100.0 & 98.2 & 100.0 & 100.0 & 94.6 & 100.0 & 98.7 & 100.0 & 100.0 & 93.8 & 91.1 & 92.9 & 67.9 & 95.2 \\
top-$k$ 128 & 100.0 & 100.0 & 100.0 & 100.0 & 96.4 & 89.3 & 95.1 & 100.0 & 100.0 & 93.2 & 89.9 & 94.6 & 67.9 & 94.3 \\
top-$k$ 512 & 100.0 & 100.0 & 100.0 & 98.2 & 100.0 & 100.0 & 99.1 & 100.0 & 100.0 & 97.1 & 86.9 & 96.4 & 69.6 & 96.0 \\
\bottomrule
\end{tabular*}
\par\medskip

\begin{tabular*}{\textwidth}{@{\extracolsep{\fill}}lrrrrrrrrrrrrrr}
\toprule
\rlap{131K}\phantom{standard RoPE + YaRN} & S1 & S2 & S3 & MK1 & MK2 & MK3 & MV & MQ & VT & CWE & FWE & QA1 & QA2 & Avg \\
\midrule
InfLLM & 100.0 & 7.1 & 0.0 & 5.4 & 57.1 & 0.0 & 7.6 & 3.6 & 60.0 & 36.4 & 86.3 & 8.9 & 8.9 & 29.3 \\
InfiniteHiP & 100.0 & 91.1 & 91.1 & 73.2 & 32.1 & 0.0 & 76.3 & 86.6 & 57.1 & 55.9 & 86.3 & 55.4 & 23.2 & 63.7 \\
Dense + YaRN & 100.0 & 100.0 & 98.2 & 94.6 & 64.3 & 41.1 & 91.5 & 94.6 & 95.7 & 70.7 & 85.7 & 58.9 & 44.6 & 80.0 \\
Dense + DCA & 100.0 & 98.2 & 100.0 & 78.6 & 73.2 & 41.1 & 81.7 & 94.6 & 90.4 & 79.3 & 90.5 & 60.7 & 32.1 & 78.5 \\
MoBA + YaRN & 87.5 & 10.7 & 25.0 & 19.6 & 14.3 & 1.8 & 18.8 & 8.9 & 81.8 & 41.8 & 58.3 & 30.4 & 16.1 & 31.9 \\
MoBA + DCA & 98.2 & 14.3 & 17.9 & 19.6 & 23.2 & 8.9 & 17.9 & 12.1 & 81.8 & 48.2 & 88.7 & 37.5 & 25.0 & 37.9 \\
DSA + YaRN & 100.0 & 100.0 & 100.0 & 94.6 & 66.1 & 26.8 & 88.8 & 96.4 & 98.2 & 80.5 & 84.5 & 67.9 & 42.9 & 80.5 \\
DSA + DCA & 100.0 & 98.2 & 100.0 & 78.6 & 69.6 & 37.5 & 79.0 & 92.9 & 89.3 & 81.8 & 86.3 & 62.5 & 32.1 & 77.5 \\
\addlinespace[2pt]
RecSA (ours) & 100.0 & 100.0 & 100.0 & 100.0 & 76.8 & 37.5 & 92.0 & 99.6 & 99.6 & 84.8 & 88.7 & 57.1 & 46.4 & \textbf{83.3} \\
\midrule
Qwen3-4B + YaRN & 100.0 & 92.9 & 100.0 & 85.7 & 50.0 & 5.4 & 75.4 & 87.9 & 57.1 & 34.6 & 80.4 & 46.4 & 28.6 & 65.0 \\
Qwen3-4B + DCA & 100.0 & 98.2 & 100.0 & 69.6 & 55.4 & 12.5 & 78.1 & 89.7 & 52.1 & 45.7 & 84.5 & 53.6 & 19.6 & 66.1 \\
\addlinespace[2pt]
w/o summaries & 100.0 & 100.0 & 100.0 & 100.0 & 71.4 & 35.7 & 91.5 & 99.1 & 99.6 & 80.9 & 87.5 & 57.1 & 41.1 & 81.8 \\
w/o MLP term & 100.0 & 94.6 & 96.4 & 92.9 & 69.6 & 32.1 & 85.7 & 92.9 & 99.3 & 79.5 & 87.5 & 64.3 & 44.6 & 80.0 \\
standard RoPE + YaRN & 100.0 & 80.4 & 92.9 & 71.4 & 32.1 & 0.0 & 72.3 & 79.9 & 96.1 & 78.2 & 83.9 & 48.2 & 25.0 & 66.2 \\
standard RoPE & 0.0 & 0.0 & 0.0 & 0.0 & 0.0 & 0.0 & 0.0 & 0.0 & 0.0 & 0.0 & 13.1 & 0.0 & 0.0 & 1.0 \\
top-$k$ 128 & 100.0 & 94.6 & 98.2 & 98.2 & 53.6 & 12.5 & 87.1 & 98.2 & 99.6 & 83.6 & 90.5 & 57.1 & 35.7 & 77.6 \\
top-$k$ 512 & 100.0 & 100.0 & 100.0 & 100.0 & 89.3 & 67.9 & 91.1 & 99.6 & 100.0 & 88.6 & 88.1 & 71.4 & 48.2 & 88.0 \\
\bottomrule
\end{tabular*}
\par\medskip

\begin{tabular*}{\textwidth}{@{\extracolsep{\fill}}lrrrrrrrrrrrrrr}
\toprule
\rlap{262K}\phantom{standard RoPE + YaRN} & S1 & S2 & S3 & MK1 & MK2 & MK3 & MV & MQ & VT & CWE & FWE & QA1 & QA2 & Avg \\
\midrule
InfLLM & 100.0 & 3.6 & 0.0 & 3.6 & 55.4 & 0.0 & 4.0 & 4.5 & 63.2 & 30.2 & 86.9 & 17.9 & 10.7 & 29.2 \\
InfiniteHiP & 100.0 & 80.4 & 85.7 & 62.5 & 28.6 & 0.0 & 62.9 & 71.4 & 72.9 & 32.9 & 86.9 & 44.6 & 21.4 & 57.7 \\
Dense + YaRN & 98.2 & 69.6 & 80.4 & 50.0 & 26.8 & 1.8 & 45.1 & 42.4 & 85.4 & 23.4 & 57.1 & 32.1 & 16.1 & 48.3 \\
Dense + DCA & 100.0 & 89.3 & 96.4 & 60.7 & 66.1 & 25.0 & 75.9 & 75.9 & 87.5 & 46.8 & 88.7 & 50.0 & 26.8 & 68.4 \\
MoBA + YaRN & 0.0 & 0.0 & 0.0 & 1.8 & 8.9 & 0.0 & 0.4 & 0.9 & 0.0 & 0.7 & 36.9 & 7.1 & 8.9 & 5.1 \\
MoBA + DCA & 94.6 & 1.8 & 1.8 & 1.8 & 8.9 & 0.0 & 2.7 & 3.1 & 69.6 & 16.6 & 85.1 & 12.5 & 10.7 & 23.8 \\
DSA + YaRN & 100.0 & 75.0 & 87.5 & 67.9 & 23.2 & 0.0 & 48.2 & 63.8 & 93.6 & 45.0 & 72.0 & 42.9 & 23.2 & 57.1 \\
DSA + DCA & 100.0 & 85.7 & 98.2 & 64.3 & 51.8 & 12.5 & 63.8 & 71.9 & 88.2 & 64.8 & 78.0 & 50.0 & 25.0 & 65.7 \\
\addlinespace[2pt]
RecSA (ours) & 100.0 & 98.2 & 96.4 & 96.4 & 51.8 & 12.5 & 80.4 & 98.2 & 99.6 & 79.3 & 87.5 & 51.8 & 33.9 & \textbf{75.9} \\
\midrule
Qwen3-4B + YaRN & 100.0 & 60.7 & 67.9 & 51.8 & 10.7 & 0.0 & 43.3 & 45.1 & 38.9 & 3.2 & 58.3 & 28.6 & 8.9 & 39.8 \\
Qwen3-4B + DCA & 100.0 & 89.3 & 100.0 & 66.1 & 35.7 & 3.6 & 64.7 & 69.2 & 54.6 & 17.9 & 82.7 & 39.3 & 28.6 & 57.8 \\
\addlinespace[2pt]
w/o summaries & 100.0 & 100.0 & 98.2 & 94.6 & 33.9 & 3.6 & 83.5 & 97.3 & 99.3 & 70.5 & 88.7 & 50.0 & 37.5 & 73.6 \\
w/o MLP term & 100.0 & 66.1 & 83.9 & 80.4 & 42.9 & 7.1 & 54.9 & 64.3 & 99.6 & 72.3 & 85.1 & 51.8 & 28.6 & 64.4 \\
standard RoPE + YaRN & 87.5 & 8.9 & 5.4 & 10.7 & 0.0 & 0.0 & 5.8 & 1.8 & 69.6 & 25.4 & 58.3 & 8.9 & 5.4 & 22.1 \\
standard RoPE & 0.0 & 0.0 & 0.0 & 0.0 & 0.0 & 0.0 & 0.0 & 0.0 & 0.0 & 0.0 & 0.0 & 0.0 & 0.0 & 0.0 \\
top-$k$ 128 & 100.0 & 87.5 & 92.9 & 94.6 & 30.4 & 1.8 & 71.4 & 90.2 & 99.6 & 74.6 & 89.3 & 42.9 & 30.4 & 69.7 \\
top-$k$ 512 & 100.0 & 100.0 & 100.0 & 94.6 & 64.3 & 14.3 & 78.1 & 98.7 & 99.3 & 84.6 & 90.5 & 58.9 & 37.5 & 78.5 \\
\bottomrule
\end{tabular*}
\par\medskip
}
\vskip -0.1in
\end{table*}

\begin{table*}[p]
\caption{RULER score per task, 524K--2M (continued from Table~\ref{tab:ruler-tasks}). The best Avg among the methods of Table~1 is in bold. $^\dagger$: the first 4 of the 20 samples of each task.}
\label{tab:ruler-tasks-2}
\vskip 0.15in
\centering
\footnotesize
{\setlength{\tabcolsep}{2pt}%
\begin{tabular*}{\textwidth}{@{\extracolsep{\fill}}lrrrrrrrrrrrrrr}
\toprule
\rlap{524K}\phantom{standard RoPE + YaRN} & S1 & S2 & S3 & MK1 & MK2 & MK3 & MV & MQ & VT & CWE & FWE & QA1 & QA2 & Avg \\
\midrule
InfLLM & 100.0 & 10.0 & 0.0 & 10.0 & 25.0 & 0.0 & 2.5 & 3.8 & 68.0 & 10.0 & 81.7 & 30.0 & 15.0 & 27.4 \\
InfiniteHiP & 95.0 & 75.0 & 80.0 & 65.0 & 10.0 & 0.0 & 55.0 & 63.8 & 50.0 & 10.5 & 83.3 & 35.0 & 20.0 & 49.4 \\
Dense + YaRN & 10.0 & 0.0 & 0.0 & 0.0 & 0.0 & 0.0 & 0.0 & 0.0 & 0.0 & 0.0 & 31.7 & 0.0 & 5.0 & 3.6 \\
Dense + DCA & 100.0 & 85.0 & 90.0 & 55.0 & 70.0 & 0.0 & 67.5 & 55.0 & 76.0 & 12.0 & 88.3 & 35.0 & 20.0 & 58.0 \\
MoBA + YaRN & 0.0 & 0.0 & 0.0 & 0.0 & 0.0 & 0.0 & 0.0 & 0.0 & 0.0 & 0.0 & 11.7 & 0.0 & 0.0 & 0.9 \\
MoBA + DCA & 85.0 & 0.0 & 5.0 & 0.0 & 0.0 & 0.0 & 0.0 & 0.0 & 38.0 & 6.0 & 76.7 & 10.0 & 15.0 & 18.1 \\
DSA + YaRN & 100.0 & 10.0 & 0.0 & 0.0 & 0.0 & 0.0 & 7.5 & 0.0 & 68.0 & 0.0 & 53.3 & 0.0 & 5.0 & 18.8 \\
DSA + DCA & 100.0 & 80.0 & 90.0 & 55.0 & 55.0 & 0.0 & 47.5 & 42.5 & 84.0 & 42.0 & 76.7 & 30.0 & 15.0 & 55.2 \\
\addlinespace[2pt]
RecSA (ours) & 100.0 & 95.0 & 95.0 & 95.0 & 45.0 & 0.0 & 68.8 & 92.5 & 100.0 & 57.0 & 83.3 & 45.0 & 30.0 & \textbf{69.7} \\
\midrule
Qwen3-4B + YaRN$^\dagger$ & 0.0 & 0.0 & 0.0 & 0.0 & 0.0 & 0.0 & 0.0 & 6.2 & 0.0 & 0.0 & 16.7 & 0.0 & 0.0 & 1.8 \\
Qwen3-4B + DCA & 100.0 & 90.0 & 100.0 & 35.0 & 15.0 & 0.0 & 47.5 & 42.5 & 61.0 & 2.0 & 81.7 & 45.0 & 20.0 & 49.2 \\
\addlinespace[2pt]
w/o summaries & 100.0 & 90.0 & 100.0 & 95.0 & 10.0 & 0.0 & 80.0 & 96.2 & 100.0 & 43.5 & 93.3 & 40.0 & 30.0 & 67.5 \\
w/o MLP term & 100.0 & 50.0 & 90.0 & 25.0 & 10.0 & 0.0 & 38.8 & 42.5 & 100.0 & 40.0 & 88.3 & 35.0 & 35.0 & 50.4 \\
standard RoPE + YaRN & 0.0 & 0.0 & 0.0 & 0.0 & 0.0 & 0.0 & 0.0 & 0.0 & 0.0 & 0.0 & 25.0 & 0.0 & 0.0 & 1.9 \\
standard RoPE & 0.0 & 0.0 & 0.0 & 0.0 & 0.0 & 0.0 & 0.0 & 0.0 & 0.0 & 0.0 & 0.0 & 0.0 & 0.0 & 0.0 \\
top-$k$ 128 & 100.0 & 90.0 & 90.0 & 90.0 & 30.0 & 0.0 & 63.8 & 85.0 & 99.0 & 60.5 & 93.3 & 45.0 & 30.0 & 67.4 \\
top-$k$ 512 & 100.0 & 95.0 & 100.0 & 95.0 & 55.0 & 0.0 & 67.5 & 93.8 & 100.0 & 72.5 & 91.7 & 45.0 & 35.0 & 73.1 \\
\bottomrule
\end{tabular*}
\par\medskip

\begin{tabular*}{\textwidth}{@{\extracolsep{\fill}}lrrrrrrrrrrrrrr}
\toprule
\rlap{1M}\phantom{standard RoPE + YaRN} & S1 & S2 & S3 & MK1 & MK2 & MK3 & MV & MQ & VT & CWE & FWE & QA1 & QA2 & Avg \\
\midrule
InfLLM & 100.0 & 6.2 & 0.0 & 0.0 & 56.2 & 0.0 & 1.6 & 3.1 & 53.8 & 7.5 & 81.2 & 18.8 & 6.2 & 25.7 \\
InfiniteHiP & 87.5 & 62.5 & 43.8 & 37.5 & 6.2 & 0.0 & 42.2 & 37.5 & 52.5 & 1.9 & 85.4 & 18.8 & 0.0 & 36.6 \\
Dense + YaRN & 0.0 & 0.0 & 0.0 & 0.0 & 0.0 & 0.0 & 0.0 & 0.0 & 0.0 & 0.0 & 20.8 & 0.0 & 0.0 & 1.6 \\
Dense + DCA & 100.0 & 56.2 & 75.0 & 56.2 & 6.2 & 0.0 & 29.7 & 26.6 & 82.5 & 2.5 & 79.2 & 6.2 & 0.0 & 40.0 \\
MoBA + YaRN & 0.0 & 0.0 & 0.0 & 0.0 & 0.0 & 0.0 & 0.0 & 0.0 & 0.0 & 0.0 & 12.5 & 0.0 & 0.0 & 1.0 \\
MoBA + DCA & 75.0 & 0.0 & 0.0 & 0.0 & 0.0 & 0.0 & 0.0 & 0.0 & 15.0 & 0.0 & 60.4 & 6.2 & 6.2 & 12.5 \\
DSA + YaRN & 43.8 & 0.0 & 0.0 & 0.0 & 0.0 & 0.0 & 0.0 & 0.0 & 1.2 & 0.0 & 27.1 & 6.2 & 0.0 & 6.0 \\
DSA + DCA & 100.0 & 31.2 & 62.5 & 43.8 & 6.2 & 0.0 & 29.7 & 20.3 & 71.2 & 21.9 & 72.9 & 6.2 & 12.5 & 36.8 \\
\addlinespace[2pt]
RecSA (ours) & 100.0 & 87.5 & 100.0 & 81.2 & 25.0 & 6.2 & 54.7 & 90.6 & 95.0 & 23.1 & 85.4 & 31.2 & 6.2 & \textbf{60.5} \\
\midrule
Qwen3-4B + DCA & 100.0 & 50.0 & 62.5 & 31.2 & 6.2 & 0.0 & 26.6 & 17.2 & 47.5 & 1.2 & 72.9 & 18.8 & 12.5 & 34.4 \\
\addlinespace[2pt]
w/o summaries & 100.0 & 87.5 & 100.0 & 87.5 & 6.2 & 0.0 & 59.4 & 93.8 & 100.0 & 11.9 & 89.6 & 25.0 & 0.0 & 58.5 \\
w/o MLP term & 100.0 & 0.0 & 62.5 & 25.0 & 12.5 & 0.0 & 28.1 & 21.9 & 96.2 & 20.0 & 85.4 & 31.2 & 0.0 & 37.1 \\
standard RoPE + YaRN & 0.0 & 0.0 & 0.0 & 0.0 & 0.0 & 0.0 & 0.0 & 0.0 & 0.0 & 0.0 & 25.0 & 0.0 & 0.0 & 1.9 \\
standard RoPE & 0.0 & 0.0 & 0.0 & 0.0 & 0.0 & 0.0 & 0.0 & 0.0 & 0.0 & 0.0 & 0.0 & 0.0 & 0.0 & 0.0 \\
top-$k$ 128 & 100.0 & 68.8 & 81.2 & 56.2 & 6.2 & 0.0 & 50.0 & 76.6 & 100.0 & 27.5 & 87.5 & 18.8 & 12.5 & 52.7 \\
top-$k$ 512 & 100.0 & 87.5 & 100.0 & 81.2 & 25.0 & 0.0 & 57.8 & 95.3 & 92.5 & 60.0 & 89.6 & 43.8 & 12.5 & 65.0 \\
\bottomrule
\end{tabular*}
\par\medskip

\begin{tabular*}{\textwidth}{@{\extracolsep{\fill}}lrrrrrrrrrrrrrr}
\toprule
\rlap{2M}\phantom{standard RoPE + YaRN} & S1 & S2 & S3 & MK1 & MK2 & MK3 & MV & MQ & VT & CWE & FWE & QA1 & QA2 & Avg \\
\midrule
InfLLM & 100.0 & 0.0 & 0.0 & 6.2 & 25.0 & 0.0 & 0.0 & 1.6 & 40.0 & 2.5 & 83.3 & 25.0 & 31.2 & 24.2 \\
InfiniteHiP & 81.2 & 37.5 & 50.0 & 25.0 & 6.2 & 0.0 & 28.1 & 15.6 & 45.0 & 6.9 & 83.3 & 25.0 & 37.5 & 34.0 \\
Dense + YaRN & 0.0 & 0.0 & 0.0 & 0.0 & 0.0 & 0.0 & 0.0 & 0.0 & 0.0 & 0.0 & 14.6 & 0.0 & 0.0 & 1.1 \\
Dense + DCA & 100.0 & 12.5 & 25.0 & 25.0 & 0.0 & 0.0 & 15.6 & 3.1 & 65.0 & 0.0 & 81.2 & 18.8 & 12.5 & 27.6 \\
MoBA + YaRN & 0.0 & 0.0 & 0.0 & 0.0 & 0.0 & 0.0 & 0.0 & 0.0 & 0.0 & 0.0 & 8.3 & 0.0 & 0.0 & 0.6 \\
MoBA + DCA & 37.5 & 0.0 & 0.0 & 0.0 & 0.0 & 0.0 & 0.0 & 0.0 & 3.8 & 0.0 & 52.1 & 0.0 & 0.0 & 7.2 \\
DSA + YaRN & 0.0 & 0.0 & 0.0 & 0.0 & 0.0 & 0.0 & 0.0 & 0.0 & 0.0 & 0.0 & 18.8 & 0.0 & 0.0 & 1.4 \\
DSA + DCA & 100.0 & 0.0 & 12.5 & 12.5 & 0.0 & 0.0 & 12.5 & 0.0 & 73.8 & 9.4 & 54.2 & 12.5 & 6.2 & 22.6 \\
\addlinespace[2pt]
RecSA (ours) & 100.0 & 75.0 & 56.2 & 68.8 & 6.2 & 0.0 & 39.1 & 70.3 & 95.0 & 5.0 & 75.0 & 25.0 & 25.0 & \textbf{49.3} \\
\bottomrule
\end{tabular*}
}
\vskip -0.1in
\end{table*}

\section{Short-Context Benchmarks, Long-Document Tasks and Perplexity}
\label{app:general}

\subsection{Short-Context Benchmarks}
\label{app:general-short}

The three models, Qwen3-4B, Dense and \recsa, receive the same questions and prompts. The generative benchmarks use the
chat template of Qwen3-4B in non-thinking mode with greedy decoding. For \recsa, the score of each sample depends only on its first
4{,}096 tokens. In a multiple-choice question, these include the input and the option being scored; in a generation
task, they include the input and the output up to the end of the answer or the code. In this range, \recsa is equivalent to dense causal attention (Section~3.1), so we evaluate the trained weights of \recsa with dense attention, using the scaled dot-product attention (SDPA) of PyTorch~\citep{paszke2019pytorch} in Transformers~\citep{wolf2020transformers}.
Qwen3-4B and Dense are evaluated the same way.

The benchmarks are evaluated as follows:
\begin{itemize}
\item MMLU~\citep{hendrycks2021mmlu}: all 14{,}042 questions, 5-shot with the first five questions of the dev set of each subject as examples,
in the format of lm-evaluation-harness~\citep{gao2024lmeval} and without the chat template. We compare the log-likelihoods of the option letters A--D.
\item ARC-C~\citep{clark2018arc}: all 1{,}172 questions, \mbox{0-shot}. As in MMLU, the prompt gives the question and its
options labeled with letters, without the chat template, and we compare the log-likelihoods of the letters.
\item GSM8K~\citep{cobbe2021gsm8k}: all 1{,}319 questions, \mbox{0-shot}, with ``Please reason step by step, and put your final answer within
\texttt{\textbackslash boxed\{\}}.'' appended to the question and at most 1{,}024 generated tokens. The answer is the number in
the last \texttt{\textbackslash boxed\{\}}, or the last number in the response if it has none.
\item IFEval~\citep{zhou2023ifeval}: 541 prompts with at most 1{,}280 generated tokens, scored by the checkers of lm-evaluation-harness 0.4.13;
we report prompt-level strict accuracy. The checkers draw random numbers, so we set the seed of Python's \texttt{random} module to 0,
of NumPy to 1234 and of langdetect to 0.
\item HumanEval~\citep{chen2021humaneval} and MBPP~\citep{austin2021mbpp}, with 164 and 500 problems: the \texttt{humaneval\_instruct} and \texttt{mbpp\_instruct} formats of lm-evaluation-harness, \mbox{0-shot},
pass@1, with at most 1{,}024 and 512 generated tokens, respectively. Each program runs separately with a time limit of
10~s.
\end{itemize}

Table~\ref{tab:general-all} lists the scores of the three models on all benchmarks, including LongBench
(Appendix~\ref{app:general-long}).

\begin{table}[!t]
\caption{Scores of the original Qwen3-4B, the Dense backbone and \recsa. Dense is trained from Qwen3-4B with the settings
of Stage~2 on the same sequences as \recsa (Section~5.1). Best results in each row are in bold.}
\label{tab:general-all}
\vskip 0.15in
\centering
\small
\setlength{\tabcolsep}{4pt}
\begin{tabular}{lrrr}
\toprule
Benchmark & Qwen3-4B & Dense & RecSA \\
\midrule
MMLU (5-shot) & 70.1 & 70.3 & \textbf{70.5} \\
ARC-C (0-shot) & 87.2 & 86.9 & \textbf{88.1} \\
GSM8K (0-shot, CoT) & \textbf{91.5} & 90.9 & 90.7 \\
IFEval (prompt, strict) & 79.1 & \textbf{79.5} & 78.6 \\
HumanEval (0-shot, pass@1) & \textbf{82.3} & 81.1 & 81.7 \\
MBPP (0-shot, pass@1) & \textbf{63.6} & 60.8 & 61.8 \\
\midrule
LongBench & 49.2 & 50.9 & \textbf{51.8} \\
\bottomrule
\end{tabular}
\vskip -0.1in
\end{table}

\subsection{Long-Document Tasks}
\label{app:general-long}

\paragraph{LongBench.} Of the 14 English tasks of LongBench~\citep{bai2024longbench}, we use 12 (NarrativeQA,
\mbox{MultiFieldQA-en}, HotpotQA, 2WikiMQA, GovReport, QMSum, MultiNews, TREC, TriviaQA, SAMSum, PassageCount and
PassageRetrieval-en), together with the 2 code tasks (LCC and \mbox{RepoBench-P}), 3{,}350 samples in total. Qasper and MuSiQue are sources of the training data
(Appendix~\ref{app:config-data}) and are excluded. The prompts, the maximum generation length of each task and the
metrics follow the official settings. As in the official code, prompts longer than 31{,}500 tokens, the limit for 32K
models, are truncated in the middle, and TREC, TriviaQA, SAMSum, LCC and \mbox{RepoBench-P} are used without the chat
template. The other tasks use the chat template of Qwen3-4B in non-thinking mode. All prompts fit within the training length of Qwen3-4B,
so Qwen3-4B runs without YaRN. All models use greedy decoding, and the score is the mean over the 14 tasks.

\paragraph{En.MC.} En.MC from InfiniteBench~\citep{zhang2024infinitebench} has 229 questions with four options on 58 novels, in which the names of the
characters are replaced. We input the whole novel without truncation, so the inputs have 70K to 746K tokens, including
the question, the options and the template. The prompt uses the template that InfiniteBench provides for chat models,
as one user turn in non-thinking mode; we generate at most 40 tokens with greedy decoding and score the output with the official method. Each
question is assigned to the smallest of 131K, 262K, 524K and 1M that is at least as long as its input, and each baseline uses its settings
at that length (Appendix~\ref{app:config-baselines}). Table~\ref{tab:enmc} lists the scores.

\begin{table*}[!t]
\caption{En.MC scores. The last four columns group the questions by input length, with the number of questions in
parentheses. Best results in each column are in bold.}
\label{tab:enmc}
\vskip 0.15in
\centering
\small
\setlength{\tabcolsep}{6pt}
\begin{tabular}{lrrrrr}
\toprule
Method & All (229) & 70K--131K (79) & 131K--262K (106) & 262K--524K (40) & 524K--746K (4) \\
\midrule
Qwen3-4B + YaRN & 34.5 & 64.6 & 20.8 & 15.0 & 0.0 \\
InfiniteHiP & 66.4 & 67.1 & \textbf{68.9} & \textbf{57.5} & \textbf{75.0} \\
Dense + DCA & 62.4 & 69.6 & 64.2 & 45.0 & 50.0 \\
DSA + DCA & 62.0 & 64.6 & 65.1 & 52.5 & 25.0 \\
\addlinespace[2pt]
RecSA (ours) & \textbf{66.8} & \textbf{74.7} & 66.0 & 55.0 & 50.0 \\
\bottomrule
\end{tabular}
\vskip -0.1in
\end{table*}

\subsection{Perplexity}
\label{app:ppl}

\paragraph{Books.} We take the books of the PG-19~\citep{rae2020pg19} training set that have at least $10^6$ tokens under the Qwen3
tokenizer and exclude two kinds. First, we exclude reference works made of unrelated entries, such as dictionaries and
encyclopedias. Second, we exclude books that overlap with the training data: their title, URL or full-text hash
appears in the metadata of the training data, or a training sequence contains one of their 50-word spans or one of their 50-token spans taken every
256 tokens. This check covers all sequences of the training pool
(Appendix~\ref{app:config-data}). Of several editions or volumes of one work, we keep one. Nine books remain, and
Table~\ref{tab:ppl-books} lists them.

\begin{table}[!t]
\caption{The nine PG-19 books. Tokens: length of the whole book.}
\label{tab:ppl-books}
\vskip 0.15in
\centering
\footnotesize
\setlength{\tabcolsep}{3pt}
\begin{tabular}{@{}>{\raggedright\arraybackslash}p{4.75cm}rr@{}}
\toprule
Title & Gutenberg ID & Tokens \\
\midrule
Modern Machine-Shop Practice, Vols.~I and II & 39225 & 1,261,044 \\
The Bible, King James Version & 30 & 1,405,554 \\
History of the United Netherlands, 1584--1609 & 4885 & 1,003,719 \\
The Anatomy of Melancholy & 10800 & 1,000,033 \\
The Complete Memoirs of Jacques Casanova de Seingalt & 2981 & 1,531,970 \\
The History of England, Vol.~II (Smollett) & 19217 & 1,099,801 \\
Abridgment of the Debates of Congress, 1789--1856 & 40499 & 1,073,145 \\
Complete Poetic and Dramatic Works of Robert Browning & 50954 & 1,348,826 \\
Church History, Vols.~1--3 (Kurtz) & 51491 & 1,279,278 \\
\bottomrule
\end{tabular}
\vskip -0.1in
\end{table}

\paragraph{Computation.} For context length $L$, the window of each book is the last $L$ tokens of its first $10^6$
tokens. The input is the plain text of the book, without a chat template or special tokens. We run one forward pass
per window and report the exponential of the mean negative log-likelihood over the predicted tokens of the nine
windows, which have the same length. The sparse methods use the token budgets of Table~1, and YaRN
and DCA use their settings at each length (Appendix~\ref{app:config-baselines}).

\section{Efficiency}
\label{app:efficiency}

\subsection{Experimental Setup}
\label{app:eff-protocol}

\paragraph{Environment.} All measurements run on one H200 GPU with batch size~1 in bf16, with PyTorch 2.7.1, CUDA
12.8, Triton 3.3.1~\citep{tillet2019triton}, FlashAttention 2.8.3~\citep{dao2024flashattention2} and Transformers 5.9.0.

\paragraph{Prefill.} Up to 1M, the input is the window of length $L$ of \textit{The Anatomy of Melancholy}
(Appendix~\ref{app:ppl}). At 2M, it is the first $10^6$ tokens of \textit{The History of England, Vol.~II} followed by
the first $10^6$ tokens of \textit{The Anatomy of Melancholy}. We measure the wall-clock time of one complete prefill after a warm-up and report the median of 3 timed
runs up to 262K and the time of one run from 524K. For \recsa, the time includes building the KV cache and the tree and, at 1M
and 2M, writing the exact tokens to host memory. Dense runs the forward pass of Transformers with FlashAttention-2,
computes the logits of the last position only and keeps no KV cache. At 2M, the intermediate results of the Dense
forward pass over the whole sequence do not fit in GPU memory, so we compute the non-attention part in chunks of
131{,}072 tokens, while attention still calls FlashAttention once per layer over the whole sequence. In the prefill of \recsa at 2M, the
two passes of level~1, which has 125{,}000 nodes, process 4 adjacent queries per thread block instead of 16
(Section~4.1).

\paragraph{Decoding.} Up to 524K, the input is a sequence of random tokens generated with a fixed seed. At these lengths, all data
reside on the GPU, and the number of candidates scored at each level and the number of entries read depend only on the
position. At 524K, the step time on a RULER sample differs from that on the random sequence by
less than 0.1\%. At 1M and 2M, the amount read from host memory at each step depends on which nodes are
selected, so we measure \recsa on two RULER samples at each length, one from MK1 and one from MQ
(Table~2). Dense reads its whole KV cache at every step. We measure Dense on the same random sequence up to
524K and on random keys and values at 1M and 2M.

Except for Dense at 1M and 2M, we first prefill the input. We then decode token by token, and each step includes the
output logits. Up to 524K, each step takes the next token of the random sequence as input.
On the RULER samples, each step takes the token that \recsa generates greedily, with the end-of-sequence token
suppressed; \recsa without the GPU cache (Section~4.3) and the two-level variant
(Appendix~\ref{app:depth}) take the same tokens. The whole decoding step is captured as one CUDA
graph. At the top level, each of the 16 segments of Section~4.2 is scored by 4 thread blocks per KV head, so 512
thread blocks compute the importance. After a warm-up, we synchronize at the end of each step and take the median step time over at least 32
steps; at 1M and 2M, the latency of \recsa is the mean of the medians of the two samples. Dense uses the Flash-Decoding kernel~\citep{dao2023flashdecoding} of FlashAttention-2 with a preallocated KV cache, and
the number of splits is that of the default heuristic of FlashAttention or 32, whichever is faster. The non-attention part of Dense (projections, MLPs and
normalization) uses the same kernels as \recsa.

\paragraph{Offloading at 1M and 2M.} \recsa keeps the exact tokens in host memory and caches the most recent exact
tokens on the GPU (Section~4.3). On the RULER samples, the cached range holds 80.1\% of the input at 1M and
38.1\% at 2M, and 89.2\% and 77.4\% of the selected level-1 nodes fall in it. Dense keeps the keys and values of the last 949{,}248 and
945{,}152 input tokens on the GPU, 95\% and 47\% of the input, which fills the available memory. The rest stays in pinned
host memory and is copied to the GPU in chunks of 16{,}384 tokens at every step: 7.0~GiB at 1M and 144.9~GiB at 2M.
FlashAttention computes attention over the part on the GPU and over each copied chunk, and the partial results are
merged by their LogSumExp; the copies and the computation run one after another on one CUDA stream. The measured
bandwidth from host to GPU is 51.6~GiB/s.

\subsection{Latency}
\label{app:eff-latency}

Table~\ref{tab:latency} lists the prefill times behind Figure~5a and the decoding latencies of Figure~1b.

\begin{table}[!t]
\caption{Prefill time (s) and decoding latency (ms per token) of \recsa and Dense. At 1M and 2M, host memory holds the
exact tokens of \recsa and, in decoding, the part of the Dense KV cache that does not fit on the GPU.}
\label{tab:latency}
\vskip 0.15in
\centering
\small
\setlength{\tabcolsep}{4pt}
\begin{tabular}{lrrrr}
\toprule
 & \multicolumn{2}{c}{Prefill (s)} & \multicolumn{2}{c}{Decoding (ms)} \\
\cmidrule(lr){2-3} \cmidrule(lr){4-5}
Context length & RecSA & Dense & RecSA & Dense \\
\midrule
32K & 1.59 & 1.46 & 4.88 & 5.29 \\
64K & 3.87 & 4.66 & 5.99 & 6.41 \\
131K & 13.5 & 16.8 & 6.04 & 8.56 \\
262K & 34.2 & 65.0 & 6.06 & 13.3 \\
524K & 79.8 & 257 & 6.15 & 23.4 \\
1M & 199 & 908 & 8.35 & 177 \\
2M & 603 & 3,672 & 10.7 & 2,945 \\
\bottomrule
\end{tabular}
\vskip -0.1in
\end{table}

\subsection{Decoding Latency Breakdown}
\label{app:eff-breakdown}

To break down a step, we prefill the input again in a new process and run decoding steps under the PyTorch profiler,
which records the time of every kernel in the CUDA graph. Up to 524K, the work of a step depends only on its position,
so we profile 8 steps after the timed steps of Appendix~\ref{app:eff-protocol}. At 1M and 2M, it also depends on the
selected nodes, so we profile the steps at the timed positions, with the same input tokens. We assign each kernel to one of the
following parts and take the mean time of each part over the profiled steps:
\begin{itemize}
\item Non-attention: the projections, the MLPs, normalization, the embedding, the output layer and the idle time
between kernels, which is the step time of \mbox{Table}~\ref{tab:latency} minus the summed kernel time.
\item Tree update: appending the key and value of the new token to level~0 and recomputing the last node of each level
(\ding{172} and \ding{173} in Figure~4b). At 1M and 2M, this writes the new key and value to host
memory and to the GPU cache and copies the children of the last level-1 node to the GPU.
\item Top level, Intermediate levels: scoring, top-$k$ selection and attention over the summaries at that level.
\item Gather: gathering the selected exact tokens from the GPU cache and from host memory (\ding{175} in
Figure~4b).
\item Exact tokens: attention over the exact tokens in the candidate list of level 0.
\item Merge: merging the partial softmax results of the segments and the levels.
\end{itemize}
The time of a kernel that serves two adjacent levels is split between them in proportion to the
time of the work of each level run alone. Table~\ref{tab:breakdown} lists the results; the 1M and 2M columns are the means over the two
RULER samples. In a Dense step at 1M and 2M, copying keys and values from host memory takes
136 of 177~ms and 2{,}823 of 2{,}945~ms.

\begin{table}[!t]
\caption{Time breakdown (ms) of one decoding step of \recsa (Figure~5b). The time of a kernel that
serves two adjacent levels is split between them in proportion to the time of the work of each level run alone. --: the part does
not occur at this length.}
\label{tab:breakdown}
\vskip 0.15in
\centering
\footnotesize
\setlength{\tabcolsep}{3pt}
\begin{tabular}{l*{6}{>{\raggedleft\arraybackslash}p{20.5pt}}}
\toprule
Context length & 32K & 131K & 262K & 524K & 1M & 2M \\
\midrule
Tree levels & 2 & 3 & 3 & 3 & 3 & 4 \\
\midrule
Non-attention & 3.26 & 3.35 & 3.35 & 3.35 & 3.41 & 3.40 \\
Tree update & 0.25 & 0.39 & 0.39 & 0.39 & 0.63 & 0.87 \\
Top level & 0.89 & 0.73 & 0.80 & 0.91 & 1.24 & 0.70 \\
Intermediate levels & -- & 1.03 & 0.98 & 0.96 & 0.98 & 2.01 \\
Gather & -- & -- & -- & -- & 1.55 & 3.07 \\
Exact tokens & 0.35 & 0.32 & 0.32 & 0.33 & 0.32 & 0.32 \\
Merge & 0.15 & 0.21 & 0.21 & 0.21 & 0.21 & 0.28 \\
\midrule
Total & 4.88 & 6.04 & 6.06 & 6.15 & 8.35 & 10.66 \\
\bottomrule
\end{tabular}
\vskip -0.1in
\end{table}

\subsection{Tree Depth}
\label{app:depth}
Up to 65{,}536 tokens, \recsa uses at most two levels. For longer contexts, Table~\ref{tab:depth} compares \recsa with a
two-level variant, which scores every level-1 node at each step. The variant is slower than \recsa at every length, from
$1.1\times$ at 131K to $1.9\times$ at 2M, because the number of level-1 nodes it scores grows linearly with the context length.

\begin{table}[!t]
\caption{Decoding latency (ms per token) of \recsa and of a two-level variant, which scores every level-1 node at each
step, measured as in Appendix~\ref{app:eff-protocol}. Up to 65{,}536 tokens, \recsa uses at most two levels. At 1M and 2M, both store the exact tokens in host memory.}
\label{tab:depth}
\vskip 0.15in
\centering
\small
{\setlength{\tabcolsep}{3.5pt}%
\begin{tabular}{lrrrrr}
\toprule
Context length & 131K & 262K & 524K & 1M & 2M \\
\midrule
RecSA tree levels & 3 & 3 & 3 & 3 & 4 \\
Level-1 nodes & 8,192 & 16,384 & 32,768 & 62,500 & 125,000 \\
\midrule
RecSA & 6.04 & 6.06 & 6.15 & 8.35 & 10.7 \\
Two-level variant & 6.37 & 7.11 & 7.83 & 11.5 & 20.5 \\
Speedup of RecSA & $1.1\times$ & $1.2\times$ & $1.3\times$ & $1.4\times$ & $1.9\times$ \\
\bottomrule
\end{tabular}
}
\vskip -0.1in
\end{table}

\section{Training and Baselines}
\label{app:config}

\subsection{Training Data}
\label{app:config-data}

Table~\ref{tab:config-data} lists the training data: task data, whose responses depend on information in the context,
and general data, which we add to preserve the general capabilities of the original model.

\begin{table}[!t]
\caption{Training data. Pool: sequences available for training. Stage~1 and Stage~2: sequences read by each run.
Length: tokens per sequence in the pool, chat template included.}
\label{tab:config-data}
\vskip 0.15in
\centering
\small
\setlength{\tabcolsep}{2pt}
\begin{tabular}{lrrrr}
\toprule
Data & Pool & Stage~1 & Stage~2 & Length \\
\midrule
\multicolumn{5}{l}{\textit{Task data}} \\
\quad Synthetic retrieval & 5,500 & 4,769 & 4,760 & 32,060--32,456 \\
\quad Document QA & 958 & 831 & 840 & 32,749--32,787 \\
\multicolumn{5}{l}{\textit{General data}} \\
\quad Books and code & 420 & 295 & 295 & 32,741--32,768 \\
\quad Instructions & 588 & 398 & 398 & 32,768 \\
\quad Long-form & 150 & 107 & 107 & 9,325--24,623 \\
\midrule
Sequences & 7,616 & 6,400 & 6,400 & \\
Tokens & 245.2M & 206.4M & 206.4M & \\
\bottomrule
\end{tabular}
\vskip -0.1in
\end{table}

\paragraph{Task data.} The synthetic retrieval tasks insert records such as ``[R248624] E407003 code V153512.'' at line
breaks into PG-19 books~\citep{rae2020pg19} or GitHub source code and ask for values in the records, for example the
value of one key among near-duplicate keys, the values of up to 12 keys in a given order, the code reached through a
chain of links or a sum of amounts. The document QA sequences hold 849 questions from MuSiQue~\citep{trivedi2022musique}
and 109 from Qasper~\citep{dasigi2021qasper}; we add distractor passages from the training split
of the same dataset to the context of each question. Each prompt specifies the
response format, such as the answer alone or the answer with the identifiers of its supporting records or passages.

\paragraph{General data.} The book and code sequences are contiguous spans of one PG-19 book or one GitHub repository.
The instruction and long-form sequences pair a document window with one instruction, and their responses were
generated by the original Qwen3-4B in non-thinking mode with greedy decoding. The instruction sequences use six instruction
types, such as summarizing a passage or answering a question that the same model generated from the passage, with two
templates per type for PG-19 books and two for GitHub code; their responses have at most 512 tokens. After generation, we prepend earlier text
of the same document so that each instruction sequence has 32{,}768 tokens. The long-form instructions, from 14 templates, ask for about 400 to 600 words on a
window of 8{,}600 to 24{,}000 tokens; only responses of 400 to 800 tokens are kept.

\paragraph{Loss masking.} Task, instruction and long-form sequences use the chat template of Qwen3-4B in non-thinking mode,
with an empty thinking block before the response; the language modeling loss covers only their response tokens, including the final
end-of-turn token. Book and code sequences have no template, and the loss covers every next-token target.

\paragraph{Order and amount.} The sequences of each kind of data are shuffled with a fixed seed, and the synthetic retrieval and
document QA sequences are then shuffled into one order. Optimizer steps 1, 9, 17, \dots{} read only general sequences, and all
other steps read task sequences. Each run reads 800 general and 5{,}600 task sequences and no sequence twice. The two
parts of Stage~1 read the same sequences; Stage~2 reads the same general sequences and shares 5{,}200 task sequences
with Stage~1. The three runs (the two parts of Stage~1 and Stage~2) process 0.62B tokens in total and read 6{,}800
distinct sequences with 219.4M tokens.

\subsection{Training Hyperparameters}
\label{app:config-hparams}

\begin{table}[!t]
\caption{Hyperparameters of \recsa training. Each run trains only the listed parameters; all other parameters are frozen.
$\mathcal{L}_{\mathrm{KL}}$: the KL terms of each part, with the query positions and weights of
Appendix~\ref{app:config-stage1}. $\mathcal{L}_{\mathrm{aux}}$: auxiliary cross-entropy. $\mathcal{L}_{\mathrm{shift}}$: shift penalty (Eq.~(7)), with weight $\lambda = 10$. LM: language modeling loss.
Learning rate: peak value. Batch size: sequences per optimizer step. Accumulation: gradient accumulation steps. Time: wall-clock time of the training loop,
including checkpointing.}
\label{tab:config-hparams}
\vskip 0.15in
\centering
\small
\setlength{\tabcolsep}{2pt}
\begin{tabular}{lccc}
\toprule
 & \multicolumn{2}{c}{Stage~1} & \\
\cmidrule(lr){2-3}
 & First part & Second part & Stage~2 \\
\midrule
Parameters & $\mathbf{w}$, $\tau$, $g$ & $W_1$, $\mathbf{b}_1$, $W_2$ & backbone \\
Count & 4,680 & 887,040 & 4.02B \\
Loss & $\mathcal{L}_{\mathrm{KL}} + 0.1\,\mathcal{L}_{\mathrm{aux}}$ & $\mathcal{L}_{\mathrm{KL}} + 10\,\mathcal{L}_{\mathrm{shift}}$ & LM \\
$k$ & 64 & 64 & 256 \\
Learning rate & $10^{-2}$ & $3 \times 10^{-3}$ & $2.5 \times 10^{-6}$ \\
Warm-up steps & 20 & 20 & 40 \\
Steps & 400 & 400 & 800 \\
Batch size & 16 & 16 & 8 \\
Accumulation & 2 & 2 & 1 \\
Time & 37 min & 39 min & 7.7 h \\
\bottomrule
\end{tabular}
\vskip -0.1in
\end{table}

Table~\ref{tab:config-hparams} lists the hyperparameters of each run. All runs use AdamW with $\beta_1 = 0.9$,
$\beta_2 = 0.95$, $\epsilon = 10^{-8}$ and no weight decay, clip the gradient to a global norm of 1.0 and run on 8
NVIDIA H200 GPUs with one sequence per GPU in each forward and backward pass. The learning
rate rises linearly during warm-up, stays at its peak and decays linearly over the last 15\% of the steps, to $1/60$
(Stage~1) or $1/120$ (Stage~2) of the peak at the last step. In Stage~2, the loss of a sequence is the mean
cross-entropy over the tokens that the loss covers, and all sequences have equal weight, so general data, read at one step in eight, carries one
eighth of the total weight. Stage~2 keeps fp32 weights and computes in bf16; Stage~1 computes the pooler and its
losses in fp32.

\subsection{Stage~1 Query Sampling and Loss Computation}
\label{app:config-stage1}

\paragraph{Tree.} Every sampled query uses the tree of its whole sequence, even when its prefix alone would need fewer levels (Section~3.1). With $k = 64$, a training sequence of length $T$ has three levels if $T > 16{,}384$ and two otherwise;
51 of the 107 long-form sequences of Stage~1 have two. On a three-level sequence, a query at position $t < 16{,}384$ has at most 64 top-level
candidates, which are all selected and expanded; its top-level KL is still computed.

\paragraph{Query positions.} For a sequence of length $T$, we draw 256 positions $t$ uniformly, with replacement, with
$1{,}024 \le t < T$; the query set $Q$ is the union of the drawn positions and the task-specific positions $X$. In a task sequence, $X$ contains the position $t_a$ that predicts the
first response token that depends on the evidence, for example the first token of a value or of a cited record or passage identifier. In
synthetic retrieval tasks, $X$ also contains the position of the first token of the question. General data has no
task-specific positions. We write $R = Q \setminus X$.

\paragraph{Losses.} For one sequence and one layer, let $\overline{\mathrm{KL}}^{(\ell)}(S)$ be the mean of
$\mathrm{KL}(\mathbf{y}^{(\ell)}_t \Vert \boldsymbol{\pi}^{(\ell)}_t)$ (Eq.~(6)) over the 8 KV heads and
the positions $t \in S$, with $\ell = \mathrm{top}$ for the top level; at level~1, the mean leaves out the pairs of a KV head
and a position whose level-1 candidates hold less than $10^{-4}$ of the dense attention mass. Let $P(S)$ be 32 positions drawn uniformly without replacement from $S$, drawn anew for each layer. The two
parts of Stage~1 minimize
\begin{align*}
\mathcal{L}_{1} &= \overline{\mathrm{KL}}^{(\mathrm{top})}(Q) + \overline{\mathrm{KL}}^{(1)}\big(P(Q)\big)
  + 0.1\,\mathcal{L}_{\mathrm{aux}}, \\
\mathcal{L}_{2} &= 0.1\,\big[\overline{\mathrm{KL}}^{(\mathrm{top})}(R) + \overline{\mathrm{KL}}^{(1)}\big(P(R)\big)\big] \\
  &\quad + \overline{\mathrm{KL}}^{(\mathrm{top})}(X) + \overline{\mathrm{KL}}^{(1)}(X) + 10\,\mathcal{L}_{\mathrm{shift}},
\end{align*}
where the level-1 terms apply only to three-level sequences and a term over an empty set is zero. In the second part,
the mean KL over the task-specific positions has ten times the weight of the mean over the other sampled positions.

The auxiliary cross-entropy $\mathcal{L}_{\mathrm{aux}}$ is the mean of $-\log \pi^{(\mathrm{top})}_{t_a, i}$ over the 8 KV heads,
where $i$ is the top-level node that contains the start of the evidence. It applies to the 982 of the
6{,}400 sequences that have a single piece of evidence starting before $t_a$. The shift penalty $\mathcal{L}_{\mathrm{shift}}$ evaluates
Eq.~(7) for each KV head, with $M_\ell$ counting every node of level $\ell$, including the partial last
node, and averages it over the 8 KV heads; the denominator is clamped to at least $10^{-12}$. The penalty covers levels 1
and 2, or level 1 on two-level sequences, and applies to every sequence.

We sum the losses of the 36 layers and average them over the 16 sequences of each optimizer
step; the loss of each layer updates only the pooler of that layer. Gradient clipping applies jointly to the poolers of
all layers.

\subsection{Baselines}
\label{app:config-baselines}

\paragraph{Training setup.} Dense, MoBA~\citep{lu2025moba} and DSA~\citep{deepseek2025v32} start from the
original Qwen3-4B and train all backbone parameters with the Stage~2 settings of Table~\ref{tab:config-hparams}, on the same
sequences in the same order. Training uses
RoPE without scaling, and Dense uses FlashAttention-2~\citep{dao2024flashattention2}. In training, MoBA and DSA attend
to at most 5{,}888 entries per query, the token budget of \recsa at 32K.

\paragraph{MoBA.} We use the official implementation in all 36 layers, with blocks of 128 tokens. Each query head
scores the complete blocks before its position by the dot product of the query with the mean key of the block, and
attends to its current block and the $k - 1$ highest-scoring blocks. MoBA adds no parameters; $k = 46$ in training.

\paragraph{DSA.} Each layer has a lightning indexer with 16 heads of dimension 128, two per GQA group, and one indexer key
per token, normalized by LayerNorm over all 128 dimensions, with RoPE on its first 64 dimensions. The indexer has 202M parameters in total, runs in fp32
and takes the normalized input of the attention layer, detached from the backbone. For each query and GQA group, DSA
attends to the $k$ tokens with the highest index scores (Eq.~(1) of DSA, summed over the two heads of the group), without
sink tokens or a local window. As in DSA, we first warm up the indexer on the frozen original model with dense
attention, using the KL divergence of Eq.~(3) of DSA to the attention distribution averaged over the 4 heads of the group.
The warm-up runs for 950 steps of 16 sequences (0.49B tokens), about 2.4 passes over the Stage~2 sequences, with AdamW,
no weight decay and a learning rate that rises over 20 steps to $10^{-3}$ and then stays constant. Each sequence
contributes 4{,}096 query positions. If at most 2{,}048 positions are covered by the language modeling loss or predict a covered token, the
query positions include all of them, and the rest are drawn uniformly without replacement; otherwise all
4{,}096 are drawn uniformly without replacement. During backbone training, the indexer continues to train with the KL divergence restricted to the selected tokens
(Eq.~(4) of DSA), at a learning rate of $5 \times 10^{-5}$ under the Stage~2 schedule; this loss does not reach the backbone.

\paragraph{Training-free methods.} InfLLM~\citep{xiao2024infllm} and InfiniteHiP~\citep{lee2025infinitehip} run on the
original Qwen3-4B with their official implementations and their own position encoding, without YaRN. InfLLM uses blocks
of 128 tokens with 4 representative tokens each.

\begin{table}[!t]
\caption{Settings of the baselines at each context length. Token budget: entries the query at the last position attends to under
\recsa (Table~1); DSA uses it as $k$. MoBA: selected blocks of 128 tokens, the current block
included. InfLLM: initial tokens, local window and retrieved blocks of 128 tokens. InfiniteHiP: sink tokens, sliding
window and selected tokens. YaRN: scaling factor $s$; --: not applied.}
\label{tab:config-budget}
\vskip 0.15in
\centering
\small
\setlength{\tabcolsep}{3pt}
\begin{tabular}{lrrrrrr}
\toprule
 & 32K & 131K & 262K & 524K & 1M & 2M \\
\midrule
Token budget & 5,888 & 8,192 & 8,704 & 9,728 & 11,575 & 11,990 \\
MoBA blocks & 46 & 64 & 68 & 76 & 90 & 93 \\
\midrule
\multicolumn{7}{l}{\textit{InfLLM}} \\
\quad Initial & 128 & 128 & 128 & 256 & 256 & 256 \\
\quad Local & 3,840 & 5,376 & 5,760 & 6,272 & 7,479 & 7,766 \\
\quad Blocks & 15 & 21 & 22 & 25 & 30 & 31 \\
\multicolumn{7}{l}{\textit{InfiniteHiP}} \\
\quad Sink & 448 & 640 & 640 & 768 & 896 & 896 \\
\quad Window & 1,792 & 2,496 & 2,688 & 2,944 & 3,575 & 3,734 \\
\quad Selected & 3,648 & 5,056 & 5,376 & 6,016 & 7,104 & 7,360 \\
\midrule
YaRN $s$ & -- & 4 & 8 & 16 & 32 & 64 \\
\bottomrule
\end{tabular}
\vskip -0.1in
\end{table}

\paragraph{Token budgets.} Table~\ref{tab:config-budget} lists the settings at each length. MoBA selects whole blocks, so its $k$ is the budget divided by
128 and rounded down; at 1M and 2M, it attends to at most 11{,}520 and 11{,}904 entries. The configuration files in the official code of
InfLLM set 128 initial tokens, a 4{,}096-token local window and 16 retrieved blocks; the 3K setting of the InfiniteHiP
paper has 256 sink tokens, a 1{,}024-token window and 2{,}048 selected tokens. We scale these parts in proportion to the token
budget and set the local or sliding window so that the parts sum to it.

\paragraph{Context extension.} YaRN~\citep{peng2024yarn} scales the RoPE frequencies by the smallest power of two $s$
with $32{,}768\,s \ge L$ for context length $L$, with $\alpha = 1$, $\beta = 32$ and the attention logits multiplied by $(0.1 \ln s + 1)^2$;
at 32K, it is not applied. DSA~+~YaRN also scales the RoPE frequencies of the indexer. DCA~\citep{an2024dca} uses chunks of 31{,}744 tokens and a local window of 1{,}024 tokens at every length, so that no relative distance exceeds 32{,}768 and keys within 1{,}024 tokens of the query keep their original distances. DCA keeps $\theta = 10^6$ without frequency
scaling and, as in its vLLM implementation~\citep{kwon2023vllm}, multiplies the attention logits by $\gamma^2$, with
$\gamma = \max\big(1, 0.1 \ln(N / 32{,}768) + 1\big)$ for an input of $N$ tokens. MoBA~+~DCA and DSA~+~DCA compute their selection
scores at the position indices of DCA. \mbox{Qwen3-4B}~+~YaRN and \mbox{Qwen3-4B}~+~DCA are the original \mbox{Qwen3-4B}, without training, with the same
settings; Appendix~\ref{app:ruler} lists their RULER scores.

\section{Ablations}
\label{app:ablation}

\begin{table*}[!t]
\caption{RULER score in the ablation experiments (average over the 13 tasks). Avg: average over the four lengths from
131K to 1M. --: not evaluated.}
\label{tab:ablation}
\vskip 0.15in
\centering
\small
\begin{tabular}{lrrrrrr}
\toprule
Variant & 32K & 131K & 262K & 524K & 1M & Avg \\
\midrule
RecSA & 94.7 & 83.3 & 75.9 & 69.7 & 60.5 & 72.3 \\
\midrule
w/o summaries & 94.5 & 81.8 & 73.6 & 67.5 & 58.5 & 70.4 \\
w/o MLP term & 94.8 & 80.0 & 64.4 & 50.4 & 37.1 & 58.0 \\
standard RoPE + YaRN & 95.2 & 66.2 & 22.1 & 1.9 & 1.9 & 23.0 \\
standard RoPE & 95.2 & 1.0 & 0.0 & 0.0 & 0.0 & 0.3 \\
top-$k$ 128 & 94.3 & 77.6 & 69.7 & 67.4 & 52.7 & 66.9 \\
\textit{token budget} & 3,968 & 4,352 & 4,864 & 5,888 & 5,980 & \\
top-$k$ 512 & 96.0 & 88.0 & 78.5 & 73.1 & 65.0 & 76.2 \\
\textit{token budget} & 9,728 & 15,872 & 16,384 & 17,408 & 19,255 & \\
\midrule
\multicolumn{7}{l}{\textit{Backbone frozen}} \\
\quad KL + shift penalty & 76.5 & 67.7 & -- & 55.5 & -- & -- \\
\quad KL & 74.7 & 63.6 & -- & 45.3 & -- & -- \\
\quad w/o MLP term & 78.9 & 65.9 & -- & 43.2 & -- & -- \\
\quad mean pooling & 73.8 & 52.2 & -- & 16.7 & -- & -- \\
\quad LM loss & 73.6 & 40.6 & -- & 8.9 & -- & -- \\
\bottomrule
\end{tabular}

\vskip -0.1in
\end{table*}

Table~\ref{tab:ablation} gives the RULER score of each variant from 32K to 1M. The variant w/o summaries is \recsa trained and evaluated without summaries, and the variant w/o MLP
term is \recsa trained and evaluated with $W_2 = 0$ in the pooler (Eq.~(5)). The variants standard RoPE and standard RoPE + YaRN are the same model,
trained like \recsa with both stages of Section~3.4, the pooler included, but with RoPE applied at
absolute token positions, and evaluated without and with YaRN beyond 32K. Each key is rotated at its absolute position
before pooling, so the pooler pools rotated keys and the summaries receive no further rotation. The variants top-$k$ 128 and
top-$k$ 512 are the \recsa model evaluated with $k = 128$ and $k = 512$ instead of 256; the \textit{token budget} row
below each gives the entries the query at the last position attends to (for $k = 256$, see Table~1). Under \textit{Backbone frozen}, the backbone is the original model, only the
pooler is trained and the variants are evaluated at 32K, 131K and 524K. KL + shift penalty is the pooler of \recsa after
Stage~1 (Section~3.4). KL sets $\lambda = 0$ in the second part of Stage~1. The variant w/o MLP
term in this block is the pooler after the first part of Stage~1, with $W_2 = 0$. LM loss trains the MLP term in the second part of Stage~1 with the
language modeling loss instead of $\mathcal{L}_{\mathrm{KL}} + \lambda \mathcal{L}_{\mathrm{shift}}$. Mean pooling has
no trainable parameters.

Averaged over the 148 samples per task from 131K to 1M, the variant w/o summaries scores 15.5 points lower than \recsa on
MK2 and 7.8 points lower on CWE, and its average over the other 11 tasks differs from that of \recsa by 0.1 points;
Appendix~\ref{app:ruler} gives its score on each task at each length. Its perplexity on the nine PG-19 books (Table~3) is 11.24, 10.88, 11.00, 11.02 and 11.07 from 32K to 1M, compared with 11.24, 10.87, 11.00, 11.08 and
11.12 for \recsa.

\end{document}